\documentclass[sigconf]{acmart}
\AtBeginDocument{%
	}

\copyrightyear{2025} 
\acmYear{2025} 
\setcopyright{acmlicensed}\acmConference[KDD '25]{Proceedings of the 31st ACM
	SIGKDD Conference on Knowledge Discovery and Data Mining V.2}{August 3--7,
	2025}{Toronto, ON, Canada}
\acmBooktitle{Proceedings of the 31st ACM SIGKDD Conference on Knowledge
	Discovery and Data Mining V.2 (KDD '25), August 3--7, 2025, Toronto, ON, Canada}
\acmDOI{10.1145/3711896.3737006}
\acmISBN{979-8-4007-1454-2/2025/08}

\usepackage{microtype}
\usepackage{graphicx}
\usepackage{subfigure}
\usepackage{makecell}

\usepackage{amssymb}

\usepackage{pgfplots}
\usepackage{hyperref}

\usepackage{amsthm}
\usepackage[capitalize,noabbrev]{cleveref}
\usepackage{amsfonts}
\usepackage{amsmath}
\usepackage{color}
\usepackage{epstopdf}
\usepackage{xspace}
\usepackage{cleveref}
\usepackage{booktabs}
\usepackage{multirow}
\usepackage{balance}
\usepackage{float}
\usepackage{xcolor}
\usepackage{url}

\newcommand{\eg}{\emph{e.g.,}\xspace}
\newcommand{\ie}{\emph{i.e.,}\xspace}
\newcommand{\stitle}[1]{\vspace{1.6ex}\noindent{\bf #1}}
\newcommand{\stab}{\rule{0pt}{8pt}\\[-1.2ex]}
\newcommand{\HFrame}{\kw{HFrame}}
\newcommand{\PTIME}{\kw{PTIME}}
\newcommand{\dual}{\kw{DualSim}}
\newcommand{\HGIN}{\kw{HGIN}}
\newcommand{\kw}[1]{{\ensuremath {\mathsf{#1}}}\xspace}
\newcommand{\etitle}[1]{\vspace{1ex}\noindent{\underline{\em #1}}}
\newcommand{\SHP}{\kw{SHP}}
\newcommand{\eetitle}[1]{\vspace{0.8ex}\noindent{{\em #1}}}
\newcommand{\myhrule}{\rule[.5pt]{\hsize}{.5pt}}
\newcommand{\mat}[2]{{\begin{tabbing}\hspace{#1}\=\+\kill #2\end{tabbing}}}
\newcommand{\IDGNN}{\kw{IDGNN}}
\newcommand{\false}{\kw{false}}
\newcommand{\true}{\kw{true}}
\newcommand{\sstab}{\rule{0pt}{8pt}\\[-1.8ex]}

\newcommand{\cdd}[1]{|\!|{#1}|\!|}
\newcommand{\eop}{\hspace*{\fill}\mbox{$\Box$}}  

\newcommand{\GNNs}{\kw{GNNs}}
\newcommand{\bcc}{\setcounter{ccc}{1}\theccc.}
\newcommand{\Then}{\mbox{\bf then}\ }
\newcommand{\icc}{\addtocounter{ccc}{1}\theccc.}
\newcounter{ccc}
\newcommand{\Return}{\mbox{\bf return}\ }
\newcommand{\kddavailabilityurl}{https://doi.org/10.5281/zenodo.15572826}

\begin{document}

	\title{Improving Subgraph Matching by 
		Combining Algorithms
		and Graph Neural Networks}

	\author{Shuyang Guo}
	\affiliation{
		\institution{SKLCCSE, Beihang University}
		\city{Beijing}
		\country{China}}
			\email{guoshuyang@buaa.edu.cn}

	\author{Wenjin Xie}
	\affiliation{%
		\institution{SKLCCSE, Beihang University}
		\city{Beijing}
		\country{China}}
	\email{xiewj@act.buaa.edu.cn}
	
	\author{Ping Lu}
	\authornote{Corresponding author}
	\affiliation{%
		\institution{SKLCCSE, Beihang University}
		\city{Beijing}
		\country{China}}
			\email{luping@buaa.edu.cn}

	\author{Ting Deng}
\affiliation{%
	\institution{SKLCCSE, Beihang University}
	\city{Beijing}
	\country{China}}
\email{dengting@buaa.edu.cn}
	
	\author{Richong Zhang}
\affiliation{%
	\institution{SKLCCSE, Beihang University}
	\city{Beijing}
	\country{China}}
\email{zhangrc@act.buaa.edu.cn}
	
	\author{Jianxin Li}
\affiliation{%
	\institution{SKLCCSE, Beihang University}
	\city{Beijing}
	\country{China}}
\email{lijx@buaa.edu.cn}
	
	\author{Xiangping Huang}
	\affiliation{%
		\institution{TravelSky Technology Limited}
		\institution{Beijing Engineering Research Centerof Civil Aviation Big Data}
		\city{Beijing}
		\country{China}}
	\email{xphuang@travelsky.com.cn}
	
\author{Zhongyi Liu}
\affiliation{%
	\institution{TravelSky Technology Limited}
	\institution{Beijing Engineering Research Centerof Civil Aviation Big Data}
	\city{Beijing}
	\country{China}}
\email{liuzy@travelsky.com.cn}
	
	\renewcommand{\shortauthors}{Shuyang Guo et al.}

\begin{abstract}
Homomorphism is an important
structure-preserving mapping between graphs.
Given a graph $G$ and a pattern $Q$, 
the subgraph homomorphism problem 
is to find a mapping $\varphi$ from $Q$ to $G$ such that 
adjacent vertices of $Q$ are mapped to adjacent vertices 
in $G$.
Unlike the subgraph isomorphic mapping 
that is injective, 
homomorphism 
allows multiple vertices in $Q$
to map to the same vertex in $G$, 
increasing complexity.
We develop \HFrame, the first GNN-based framework 
for subgraph homomorphism,
by combining algorithms and machine learning.
We show that 
\HFrame is more expressive than the vanilla GNN,
\ie~\HFrame can distinguish more graph pairs
$(Q, G)$ such that 
$Q$ is not homomorphic to $G$.
Moreover, 
we provide a generalization error bound for \HFrame.
Using real-life and synthetic graphs, we 
show that \HFrame is up to 
101.91$\times$ faster than exact matching
algorithms,
and its average accuracy can reach 0.962.
\looseness = -1
\end{abstract}

\ccsdesc[500]{Computing methodologies~Neural networks}
\ccsdesc[500]{Computing methodologies~Model development and analysis}

\keywords{Graph Neural Network, Graph Matching, Expressive Power, Generalization Bound}

\maketitle

\ifdefempty{\kddavailabilityurl}{}{
  \begingroup
  \small\noindent\raggedright
  \textbf{KDD Availability Link:}\\
  The source code of this paper has been made publicly available at \url{\kddavailabilityurl}.
  \endgroup
}

\section{Introduction}
\label{sec-intro}
Pattern matching is 
to identify instances of a pattern $Q$ 
in a graph $G$,  and has received 
extensive interest across various fields,
\eg 
discovering motifs in biology
~\cite{chen2019hogmmnc, licheri2021grapes, tian2007saga}, 
finding experts in social networks
~\cite{shi2017multi, fei2013expfinder, brynielsson2010detecting}, 
and reasoning on knowledge graph
~\cite{xu2019cross, teru2020inductive, li2022hismatch}.
\looseness - 01

There exist two semantics for pattern matching, 
namely subgraph isomorphism and homomorphism.
(1) The isomorphic semantics demands that
the matched instances of $Q$  in 
$G$ (\ie a subgraph of $G$) must be the same as $Q$.
This semantics has been widely studied, 
\eg discovering motifs from biological network~\cite{chen2019hogmmnc, licheri2021grapes, tian2007saga}.
(2) Homomorphic semantics,
on the other hand, permits that 
$Q$ can be ``embedded'' into the 
identified  subgraph of $G$, 
\ie the  
 subgraph can be deviated from $Q$.
It has been widely used in graph query languages 
like Cypher~\cite{Cypher} and SPARQL~\cite{SPARQL}.
The difference between isomorphism and homomorphism
lies in whether different vertices in $Q$
can be mapped to the same vertex in $G$ (see Section~\ref{sec-pre}).
\looseness = -1

The graph pattern matching problem under 
isomorphic semantics
has been extensively studied,
leading to the development
of many algorithms (\eg~\cite{DpISO,RM,CECI,VEQ,Iso-match-2,Iso-match-3,Iso-match-4,Iso-match-5,Iso-match-6,Iso-match-7,Iso-match-8}; see ~\cite{Survey-matching} for a survey and comparison).
All such algorithms work in the following steps:
(1) establish a matching order ${\mathcal O}=(u_1, \ldots, u_n)$
for vertices in pattern $Q$; 
(2) compute a set ${\mathcal C}(u)$ of candidate matches 
for each pattern vertex $u$ in $Q$, \eg ${\mathcal C}(u)$ consists of all vertices in $G$ that have the same
	label with $u$;
(3) iteratively identify matches using ${\mathcal C}(u)$ as follows:
(a) select a candidate $v_1$ in ${\mathcal C}(u_1)$ 
for the first vertex $u_1$ in the order ${\mathcal O}$;
(b) from the neighbors  $u_2$ of $u_1$, 
find a candidate $v_2$ in ${\mathcal C}(u_2)$
that has the same edge labels as $u_2$
in $Q$; 
(c) continue these steps for all other
vertices in $Q$, and check whether the selected candidates
form a match of $Q$;
if not, backtrack to re-examine other vertices in ${\mathcal C}(u_2)$.
These techniques can be adapted 
 to homomorphic semantics, 
except that the identified 
 subgraphs of $G$ are not required to 
be the same as $Q$ in the last step.
However, such algorithms take exponential
runtime due to backtracking, 
and cannot efficiently handle large-scale graphs.
To address this, various optimization strategies
have been developed, including 
index structures to filter candidates~\cite{VEQ,CECI,DpISO},
and dual simulation~\cite{StrongSimulation,Simulation-2} and 
1-WL algorithms~\cite{1WL,kWL}
to approximate results.
Despite these efforts,
the algorithms  still suffer from taking long time due to 
the high complexity.
\looseness = -1

Machine learning (ML), particularly Graph Neural Networks (GNNs), 
has been applied in the subgraph matching 
problem~\cite{NeuroMatch,Sub-ML-1,Sub-ML-2,Sub-ML-3,Sub-ML-4}.
Such solutions work as follows:
(1) exploit a GNN 
to compute the embedding for each vertex 
in $Q$ and $G$, capturing the surrounding 
topological structures of the vertex;
and (2) assessing subgraph matching 
by comparing these embeddings 
within the order-embedding space~\cite{NeuroMatch}.
These algorithms are efficient, since 
they do not depend on backtracking.
However, their predictions are not accurate,
since  their expressive power  
is bounded by the 1-WL 
algorithm~\cite{xu2018powerful}, an approximate algorithm.
Moreover, these models cannot directly 
handle homomorphic semantics,
since they do not 
account for properties of homomorphic mappings,
\eg two  distinct pattern vertices 
can be mapped to the same vertex in $G$, 
allowing two graphs 
with different sizes be homomorphic to each other.
\looseness = -1

Generalization error is an important property of machine learning 
models, reflecting how well they 
work  on unseen inputs. 
Metrics for generalization error
include Rademacher complexity~\cite{Gen-bound} and
VC dimension~\cite{VC-GNN}.
The Rademacher complexity of GNN models
is established in~\cite{Gen-bound}.
However, non such bound exists 
for subgraph isomorphic or 
homomorphic mapping models. 
\looseness = -1

In this paper,  
we aim to answer the following questions.
(1) Can we integrate  algorithmic solution 
with an ML-based techniques, to
develop approximate framework for 
graph pattern matching under homomorphic semantics? 
(2) Does the proposed framework is more expressive than 
existing approximate solutions?
And (3) what is the generalization error bound of the proposed framework?
\looseness = -1

\stitle{Contribution}. Our contributions
are listed as follows.

\stab
(1) We introduce \HFrame,  a framework designed
for the subgraph 
 homomorphism problem.
Given $Q$ and $G$, it first leverages a
\PTIME algorithm \dual\cite{StrongSimulation}  for graph dual simuation, to identify candidates 
set ${\mathcal C}(u)$
for each vertex $u$ in $Q$. Then it invokes
an ML 
model \HGIN to make predictions
on these candidates.
By combining algorithmic solutions and ML techniques, 
\HGIN can deliver high-quality predictions,
striking a balance between complexity and 
accuracy.
\looseness=-1

\stab 
(2) We develop an ML 
model \HGIN for 
subgraph homomorphism in \HFrame,
building upon \kw{IDGNN}~\cite{you2021identity},
which is tailored for subgraph isomorphism.
Instead of using multisets to gather
messages as in \kw{IDGNN},
we adopt sets to gather 
embeddings and remove duplicate messages,
since two vertices can be mapped to 
the same vertex in homomorphism. 
We encode both directions and labels
of edges into embeddings 
of vertices, to improve the accuracy 
of \HGIN.
\looseness = -1

\stab
(3) We first show that \HFrame 
is more expressive than existing models,
and capable of distinguishing more non-homomorphic graphs. 
Then we provide the first generalization error bound
for models addressing the subgraph homomorphism problem.
\looseness = -1

\stab
(4) We conducted
extensive experiments on real-life and synthetic graphs,
and found the  following:
(a) \HFrame is efficient.
It on average outperforms 
existing algorithms
for subgraph homomorphism problems
by up to 101.91$\times$, and its accuracy on average is 0.962.
(b) \HFrame is effective.
Its accuracy is at least 0.921 on all graphs,
and is on average 21\% higher than existing ML solutions. 
(c) \HFrame performs well on unseen data. 
After \HFrame is trained on synthetic graphs,
its accuracy can reach 0.892, 0.874 and 0.872
on real-life graphs Citeseerx, IMDB and DBpedia, respectively.

\stitle{Related work}. We categorized the related work as follows.

\etitle{Graph pattern matching algorithm}.
Algorithms for the subgraph isomorphism 
problem have been
extensively studied.
Ullmann proposed the first backtracking-based algorithm for subgraph isomorphism~\cite{Iso-match-6}. 
Subsequent research in algorithms for exact graph pattern 
matching focuses on developing pruning methods~\cite{carletti2017introducing,carletti2019vf3}
and improving index structures~\cite{CECI,sun2020subgraph,yan2004graph,zhao2007graph}, and so on. 
\looseness=-1

Isomorphic mappings suffer from two major drawbacks:
(1) the decision version of subgraph isomorphism is 
NP-complete~\cite{cook2023complexity}, and
(2) the requirement of  
 injective mapping 
could be too strict in many cases,
especially when the data source is noisy~\cite{yuan2011efficient}. 
To address these challenges, 
(1) a series of studies focus on designing similarity functions and 
deriving matching results through  
top-$k$ candidates or the establishment of a similarity threshold
\cite{fan2010graph,cheng2013top,dutta2017neighbor,wu2013ontology,khan2013nema,zhao2013partition}. 
(2) The other studies relax the  
 injective map to less restrictive  
relations.
Bounded simulation 
is developed to bound the lengths of 
matching paths~\cite{fan2010grapha}.
 Dual simulation and 
strong simulation  are proposed
in \cite{ma2011capturing}  to 
bound directions of edges 
	and 
distances of vertices in a match, 
	respectively. 
Subsequently, various simulation-based
matching algorithms have been developed~\cite{fan2013incremental,fan2014distributed, fard2014distributed,gao2016prs,du2018personalized}.
\looseness = -1

\etitle{ML Models for graph pattern matching}.
Multiple ML models are developed for graph pattern matching~\cite{NeuroMatch,Sub-ML-1,Sub-ML-2,Sub-ML-3,Sub-ML-4}.
Most of these models are based on Graph Neural Networks (GNNs),
to compute embeddings of vertices in a graph~\cite{scarselli2008graph},
and make a prediction using the order-embedding space~\cite{NeuroMatch}
or MLP~\cite{Sub-ML-1}.
Most work focus on 
enhancing GNNs architecture~\cite{wang2022twin,wang2022equivariant,zhang2021eigen,bouritsas2022improving,feng2022powerful},
to learn more topological information. 
Zhang et al. \cite{zhang2023expressive} 
encoded distance information into vertex embeddings. 
Geerts et al. \cite{geerts2022expressiveness} 
proposed a model to aggregate information 
from $k$-order neighbors using computational operators. 
You et al. \cite{you2021identity} injected a unique 
color identity to nodes, 
and performed heterogeneous message passing 
to obtain the embedding. 
Xu et al. \cite{xu2018powerful} 
optimized the aggregation function 
by using multisets, 
to generate distinct embeddings for different nodes. 
Maron et al. \cite{maron2018invariant} 
proposed a model consisting of 
both equivariant linear layers and nonlinear activation layers. 

There has also been work on
studying properties of GNNs.
(1) The expressive power of these GNN
models can be bounded by 
the Weisfeiler-Lehman (WL) hierarchy~\cite{xu2018powerful,morris2019weisfeiler,barcelo2022weisfeiler,zhang2023expressive,GNN-WL-C},
descriptive complexity~\cite{Express-LICS},
fragments of FO logic~\cite{Expressive-FOC2}
and global feature map transformer~\cite{Express-GFMT};
see~\cite{zhang2023expressive} for a recent survey.
(2) The generalization error
bound for GNNs can be measured by 
Rademacher complexity~\cite{Gen-bound}, VC dimension~\cite{VC-GNN}
and generalization gap~\cite{Generalization-gap,Generalization-gap}.

\section{Preliminaries}
\label{sec-pre}
Let $\Gamma$ be a countably 
infinite set of labels.

\stitle{Graphs}. 
Consider directed labeled graphs $G=(V, E, L)$,
where (1) $V$ is a finite set of vertices;
(2) $E\subseteq V\times \Gamma \times V $ is a finite set 
of edges, 
where $\langle v_1, r, v_2\rangle$ is an edge
from $v_1$ to $v_2$,  
carrying label $r\in \Gamma$;
and (3) $L$ is a labeling function that 
assigns a label $L(v) \in \Gamma$
to each vertex $v$ in $V$.
Let $R_G$ be the set of all edge labels in $G$.
\looseness=-1

For each vertex $u \in V$, 
denote by $N^+_r(u)$ (resp. $N^-_r(u)$) the set of 
its   {\em outgoing} (resp. {\em incoming}) neighbors $u_1$ of $u$
that have an edge labeled $r$ from $u$ to $u_1$
(resp.  
from $u_1$ to $u$).
That is, $N^+_r(u) {=} \{u_1\mid \langle u, r, u_1\rangle {\in} E\}$
and $N^-_r(u) {=} \{u_1\mid \langle u_1, r, u\rangle {\in} E\}$.
Denote by $N(u)$ the set of all neighbors of $u$, 
\ie $N(u) {=} \bigcup_{r\in R_G} (N^-_r(u) \cup N^+_r(u))$.

\stitle{Isomorphism (Iso) and homomorphism (Hom).} 
We introduce  the semantics of 
subgraph
isomorphism and homomorphism.
 Consider a pattern  $Q{=} (V_Q, E_Q, L_Q)$ and a graph $G {=} (V_G, E_G, L_G)$.
Here, a pattern is also a directed labeled graph. 
\looseness=-1

\etitle{Subgraph isomorphism}.
An \textit{isomorphic mapping} from  
$Q$ to $G$,
denoted by 
 $Q \sqsubseteq G$ is an injective mapping $\varphi$
from $V_Q$ to $V_G$  
such that
(1) for each $u{\in} V_Q$, $u$ and $\varphi(u)$ 
carry the same label, \ie $L_Q(u) = L_G(\varphi(u))$; and
(2) the adjacency relation preserves,  
\ie~$\langle u, r, u'\rangle \in E_Q$ if and only 
	if $\langle \varphi(u), r, \varphi(u')\rangle \in E_G$. 
When such an injective mapping $\varphi$ exists, 
we say that $Q$ is subgraph isomorphic to $G$.

\etitle{Subgraph homomorphism}.
A \textit{homomorphic mapping} from $Q$ to $G$, denoted by $Q\triangleright G$,
is a mapping 
$\varphi$ from $V_Q$ to $V_G$ such that 
(1) for each $u\in V_Q$, $u$ and $\varphi(u)$ 
carry the same label, \ie $L_Q(u) = L_G(\varphi(u))$;
and (2) when there is
an edge $\langle u, r, u'\rangle $ in $E_Q$,
edge 
$\langle \varphi(u), r, \varphi(u')\rangle $
exists in $ E_G$.
Observe that (a) $\varphi$ may not be injective;
	and (b) when $\langle \varphi(u), r, \varphi(u')\rangle \in E_G$,
	$\langle u, r, u'\rangle $ may not exist in $ E_Q$.
When such a mapping $\varphi$ exists, 
we say that $Q$ is subgraph homomorphic to $G$. 

\begin{figure}
	\centerline{\includegraphics[width=0.7\columnwidth]{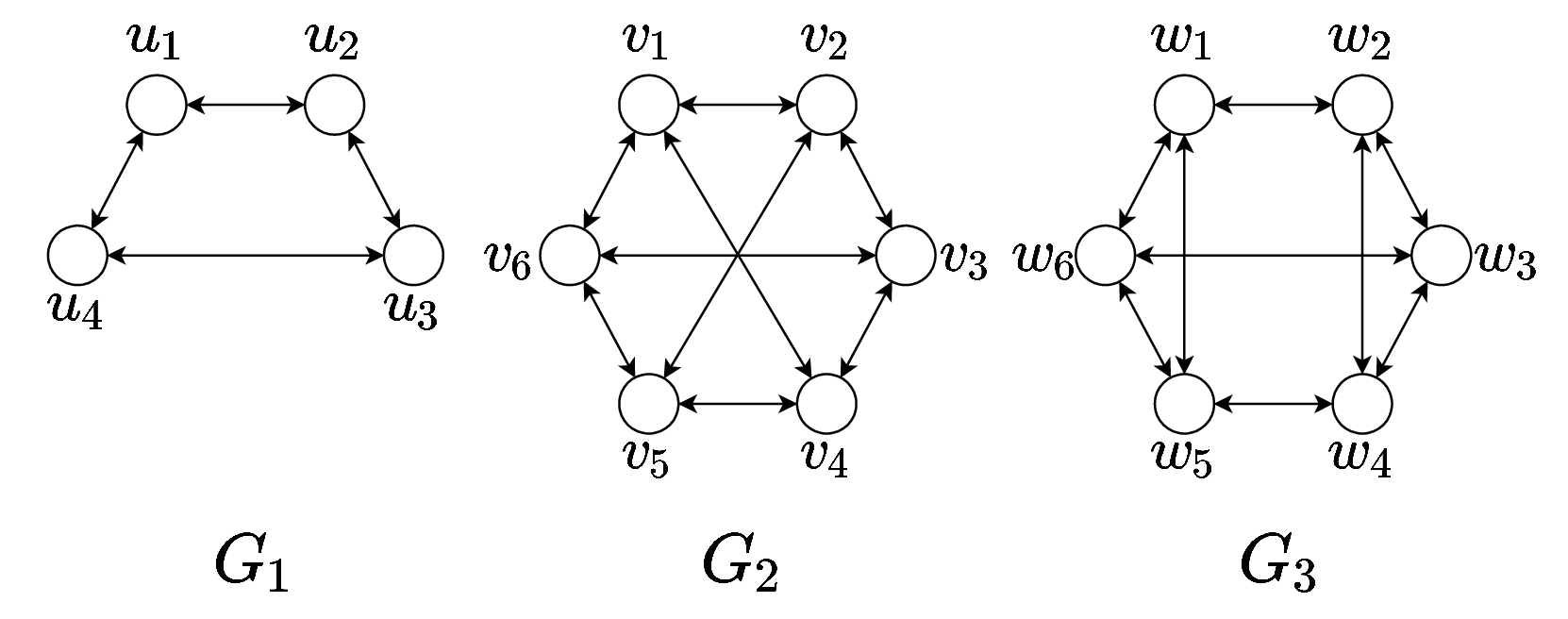}}
	\centering
	\vspace{-1.4ex}
	\caption{Differences between Iso and Hom 
	} 
	\label{fig:difference}
	\vspace{-3ex}
\end{figure}

\etitle{Differences}.
Since subgraph isomorphism 
requires that the mapping $\varphi$ is injective, 
it has the anti-symmetry property,
\ie for two graphs $G$ and $H$,
if $G{\sqsubseteq} H$ and $H{\sqsubseteq} G$, 
then $G$ and $H$ are the same graph. 
However, subgraph homomorphism does not have 
this property.
That is, there exist two distinct
graphs $G$ and $H$ such that $G\triangleright H$ and $H\triangleright G$.
\looseness=-1

\begin{example}
	Consider graphs $G_1$ and $G_2$ in Figure~\ref{fig:difference}.
	Assume that all vertices (resp.edges) in $G_1$ and $G_2$ 
	carry the same label.
	Observe that
	(a)~$G_1$ is subgraph homomorphic and subgraph isomorphic 
	to $G_2$ via mapping 
	$\varphi_1(u_i)=v_i$ $(i{=}1, 2, 3, 4)$;
	(b)~$G_2$ is not subgraph isomorphic to $G_1$,
	since $G_2$ has more vertices than $G_1$;
	and
	(c) $G_2$ is subgraph homomorphic to $G_1$ 
	via mapping 
	$\varphi_2(v_i)=u_i$ ($i{=}1,2,3, 4$),
	$\varphi_2(v_5)=u_3$ and $\varphi_2(v_6){=}u_2$. 
	That is, $G_1$ and 
	$G_2$ are subgraph homomorphic to 
	each other, but they have different numbers
	of vertices.
	\looseness=-1
	\label{exam-diff}
\end{example}

Due to such differences, solutions
for subgraph ismomorphism and homomorphism
are distinct (see Section~\ref{sec-model}).
\looseness = -1

\stitle{Problem definition}.
The paper studies the following 
subgraph 
homomorphism problem, denoted by \SHP:
{\em given a pattern $Q$ and a graph $G$, 
	whether there exists
	a homomorphic mapping $\varphi$ from $Q$ to $G$.}
\looseness=-1

\etitle{Remark}.	
Such decision problem can be applied in the following scenarios:
	(1) frequent pattern mining~\cite{ying2024representation} and graph pattern dependency discovery~\cite{Hercules}, where 
	pattern quality is usually determined by the number of vertices matching the pivot 
	(\ie a designated vertex) of $Q$, 
	which can be computed by determining 
	{\em the existence of matches} on these vertices,
	rather than enumerating the matches; 
	(2) querying for key vertices satisfying some properties defined by patterns, 
	\eg a SPARQL query from Wikidata~\cite{Sparqlexample} to find articles on Punjabi Wikipedia about Pakistani actresses, 
	which can be modeled as a pattern with Articles as the pivot;
	and (3) accelerating enumerations
	by filtering candidates
	with the decision problem
	and inducing a smaller subgraph before enumeration
	(see Section~\ref{exp:sec-exp}).
	\looseness = -1

\stitle{Existing solutions}.
In this paper, we develop an approximate solution
for \SHP.
There exist two categories of approximate solutions,
\ie algorithmic approximations and machine learning based 
solutions.
We aim to combine these two solutions to develop 
a more accurate solution.
We first 
present the two existing 
solutions.

\etitle{Dual Simulation.} 
Given a pattern $Q= (V_Q, E_Q, L_Q)$ and a graph $G= (V_G, E_G, L_G)$, 
it is to compute a matching relation $S_{(Q, G)}$ of 
vertex pairs $(u, v)$ where $u\in V_Q$ and $v\in V_G$, such that 
$u$ and $v$ have the same neighbors.
Using $S_{(Q, G)}$, 
we can approximate the subgraph homomorphic 
problem by 
checking whether  $(u, v)\in S_{(Q, G)}$.
\looseness = -1

The relation $S_{(Q, G)} \subseteq V_Q \times V_G$ 
is defined as follows: (1) for each $(u, v) \in S_{(Q, G)} $, $L_Q(u) = L_G(v)$; 
and (2) for each $u \in V_Q$, 
there exists $v \in V_G$ such that
(i) $(u, v) \in S_{(Q, G)} $; 
(ii) for each outgoing edge $(u, u_1)$ in $E_Q$, 
there is an outgoing edge $(v, v_1)$ in $E_G$ with $(u_1, v_1) \in S_{(Q, G)} $;
and (iii) for each incoming 
edge $(u_2, u)$ in $E_Q$, there is 
an incoming edge $(v_2, v)$ in $E_G$ with $(u_2, v_2) \in S_{(Q, G)} $.

The relation $S_{(Q, G)}$  can be computed in $O((|V_Q|+|E_Q|)(|V_G|+|E_G|))$ 
time via algorithm \dual~\cite{StrongSimulation} presented in Figure~\ref{alg:DualSim}.
Intuitively, 
for each vertex $u$ in $Q$, 
it first initializes the set ${\mathcal C}(u)$ 
with vertices in $G$ 
carrying the same label as $u$ (lines 1);
then it iteratively removes a candidate 
$v$ from ${\mathcal C}(u)$,
if $v$ does not have the same incoming edges
(lines 3-5) or the same outgoing 
edges  (lines 6-8)
as vertex $u$ in $Q$. 
It repeats these steps for $T$ times (line 2). 

\eetitle{Remarks}.
The number $T$ 
is bounded by $O((|V_Q|+|E_Q|)(|V_G|+|E_G|))$~\cite{StrongSimulation}.
However, only a few fixed iterations
suffice to filter candidates~\cite{DpISO}.
We set $T=2$ in the experiments (see Section~\ref{exp:sec-exp}).
\looseness = -1

\begin{figure}[tb!]
	\begin{center}
		{\small
			\begin{minipage}{3.36in}
				\myhrule
				\vspace{-1ex}
				\mat{0ex} {
					\noindent {\sl Input:\/} \hspace{1ex} \= 
					A graph $G$, a pattern $Q$ and 
					a number $T$.\\
					{\sl Output:\/} \=
					The maximum match relation $S_{(Q, G)}$ in $G$ for $Q$. \\
					\noindent
					\bcc \hspace{2ex} set 
					${\mathcal C}(u){:=}\{v \,|\, v \in V_G \land L_Q(u) = L_G(v)\}$ for each $u$ in $V_Q$;\\
					\icc \hspace{2ex} {\mbox{\bf for each}\ } $i\in [1, T]$  {\bf do} \\ 
					\icc \hspace{2ex}\hspace{2ex}{\mbox{\bf for each}\ }  $e=(u, u')\in E_Q$ and $v \in {\mathcal C}(u)$\  
					{\bf do} \\
					\icc \hspace{2ex}\hspace{2ex}\hspace{2ex}
					{\mbox{\bf if}} no edge $(v, v')$ in $G$ with $v' \in {\mathcal C}(u')$\ 
					\Then\\
					\icc \hspace{2ex}\hspace{2ex}\hspace{2ex}\hspace{2ex}${\mathcal C}(u) := {\mathcal C}(u) \setminus \{v\}$;\\
					\icc \hspace{2ex}\hspace{2ex}{\mbox{\bf for each}\ }  $e=(u', u)\in E_Q$ and $v \in {\mathcal C}(u)$\ 
					{\bf do} \\
					\icc \hspace{2ex}\hspace{2ex}\hspace{2ex}
					{\mbox{\bf if}} no edge $(v', v)$ in $G$ with $v' \in {\mathcal C}(u')$\ 
					\Then\\
					\icc \hspace{2ex}\hspace{2ex}\hspace{2ex}\hspace{2ex}${\mathcal C}(u) := {\mathcal C} \setminus \{v\}$;\\
					\icc \hspace{2ex} $S_{(Q, G)} := \{(u, v) \,|\, u \in V_Q, v \in {\mathcal C}(u)\}$;\\ 
					\icc \hspace{1ex} \Return $S_{(Q, G)}$;
				}\vspace{-1.7ex}
				\myhrule
			\end{minipage}
		}
	\end{center}
	\vspace{-2.7ex}
	\caption{Algorithm DualSim}
	\label{alg:DualSim}
	\vspace{-4ex}
\end{figure}

\etitle{Graph Neural Networks (GNNs)}.
Given a graph $G$, a pattern $Q$ and
two vertices $u$ and $v$
from $Q$ and $G$, respectively,
GNN-based solution 
checks the existence of homomorphic mapping 
as follows~\cite{Order-embedding-1,NeuroMatch}:
(1) compute the embeddings of $u$ and $v$
using GNNs; let $h_u=(a_1, \ldots, a_d)$ and
$h_v=(b_1, \ldots, b_d)$ be the embeddings
of $u$ and $v$, respectively;
and (2) exploit the order embedding space
to check the existence of homomorphic mappings,
\ie a homomorphic mapping $\varphi$ 
exists if and only if 
$a_i\leq b_i$ for all $i\in [1, d]$.
\looseness = -1

To compute the embeddings of vertices, 
we can adopt the identity-aware GNN
(\IDGNN)~\cite{you2021identity},
since \IDGNN is more expressive than vanilla GNN~\cite{you2021identity}.
Given a graph $G$ and a vertex $u$, \IDGNN computes 
a $d$-dimensional vector to represent $u$ in
$G$ by iteratively aggregating information
collected from its neighbors. 
More specifically, 
(1) \IDGNN first constructs the $k$-hop ego network $G_u$
centered at $u$, 
\ie the subgraph induced by the $k$-hop neighbors of $u$;
(2) next it computes the intial embedding of
of each vertex $v$ in $G_u$ using 
features of $v$, \eg the label of $v$;
(3) then it iteratively updates embeddings of 
all vertices in $G_u$
as follows:
(a) collect embeddings of neighbors of
a vertex $v$
in $G_u$; and (b) transform the received embeddings
via message-passing functions;
that is, let $v_1, \ldots, v_n$ be the 
neighbors of vertex $v$; the message-passing function 
for the embedding of $v_i$ is $\kw{MSG}_1$ 
if $v_i = u$; otherwise the message-passing function 
is $\kw{MSG}_0$. This step is repeated for $m$ times.
Note that (i) the ego network $G_u$ is used to 
compute the embedding of center $u$ only;
and (ii) parameter $m$ is also called the number of the layers
in \IDGNN.
\looseness = -1

\eetitle{Remark}.
These two approximate solutions have their pros and cons.

\stab
(1) \dual can process any given patterns and graphs, 
no matter whether they have been inspected before or not;
moreover, it ensures that all matches can be returned,
\ie the recall of \dual is 1;
but it may return  extra results, 
\ie it will return false positive answer.
Consider graphs $G_2$ and $G_3$ in Figure~\ref{fig:difference}.
There does not exist a homomorphic mapping 
from $G_3$ to $G_2$, since $G_3$ contains
triangles, but $G_2$ does not.
However, $S_{(G_2, G_3)} = V_{G_2} \times V_{G_3}$,
\ie $(v_i, w_j)\in S_{(Q, G)}$, $i, j\in[1, 6]$.
See the proof of Theorem~\ref{thm-hgin-dual}
for more details. 
\looseness=-1

\stab
(2) GNN models can 
achieve good accuracy in 
performing tasks like node classification~\cite{GNN-NC}, 
recommendation system~\cite{GNN-recom} and anomaly detection~\cite{GNN-AD},
and can become more expressive
using more information in the graphs~\cite{Sub-ML-1,DMPNN}. 
Moreover, the computation of GNNs
are efficient, since it only exploits
neighbors of a vertex to make a 
prediction. The computation can be further accelerated
using sampling techniques when processing large-scale
graphs~\cite{GraphSAGE}.
But GNN models may return error prediction, 
and miss some matches.

\section{Framework for homomorphic mapping}
\label{sec-model}

\begin{figure*}[th!]
	\begin{center}
		\begin{minipage}[t]{\textwidth}
			\centering
			\subfigure{\label{fig-AER}
				\includegraphics[width=0.85\columnwidth]{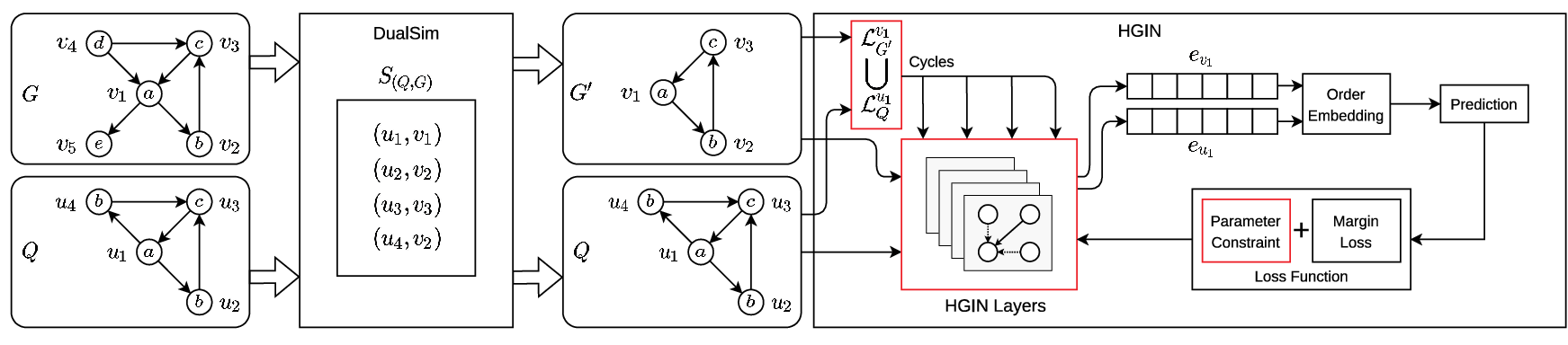}
			}
		\end{minipage}
	\end{center}
	\vspace{-3ex}
	\caption{The framework of \HFrame.
		(1) It first computes $S_{(Q, G)}$ via \dual,
		to filter candidate sets ${\mathcal C}(u)$
		for each vertex $u$ in $Q$, 
		and construct a subgraph $G'$ by removing irrelevant vertices and edges 
		from $G$;
		(2) it then generates embeddings of vertices via \HGIN; and
		(3) it finally makes a prediction via the order embedding space.
		Its novelties (marked red) consist 
		of (a) adopting sets to collect messages,
		(b) using cycles to select message-passing functions,
		and (c) integrating parameters into the loss function.
		\looseness  =-1
	}
	\label{fig:workflow}
		\vspace{-2ex}
\end{figure*}

We provide \HFrame, a framework for the subgraph homomorphic problem,
by integrating algorithms and machine learning models.
The framework is outlined in Figure~\ref{fig:workflow}.
Intuitively, the combination of \dual and \HGIN 
in \HFrame enhances the efficiency 
and accuracy of the predictions due to the following mechanisms.
(1) \kw{DualSim} removes 
 irrelevant information from 
graph $G$, which 
boosts the performance of \HGIN by focusing 
only on essential data.
(2) \HGIN optimizes the prediction by 
computing only embeddings of vertices, 
rather than embeddings of patterns and graphs.
In this way, \HGIN can leverage
the locality of subgraph matching,
\ie~\HGIN only inspects 
a small neighborhood of each vertex,
which reduces the complexity of \HGIN
and improves the accuracy of \HFrame.
(3) \HGIN only verifies candidates 
of {\em the pivot} $u_p$ of $Q$, 
since when $Q$ is homomorphic to $G$, 
the pivot $u_p$ must be mapped to some vertex in $G$.
The selection of pivot $u_p$ can be based on the degrees of vertices, 
\eg picking the vertex with the maximum degree in $Q$~\cite{Sub-ML-3}.
\looseness = -1

\stitle{Homomorphic Neural Node Matching (HGIN)}.
It is to decide, given a  patter $Q$, a graph $G$, vertex $u$ in $Q$ and vertex $v$ in $G$, 
whether there is a homomorphic mapping $\varphi$ from
$Q$ to $G$ such that $\varphi(u)=v$.
\looseness = -1

\etitle{Overview}.
We design \HGIN by extending \IDGNN~\cite{you2021identity},
to address the following three challenges from homomorphic semantics.
\looseness=-1

\begin{figure}
	\centerline{\includegraphics[width=0.7\columnwidth]{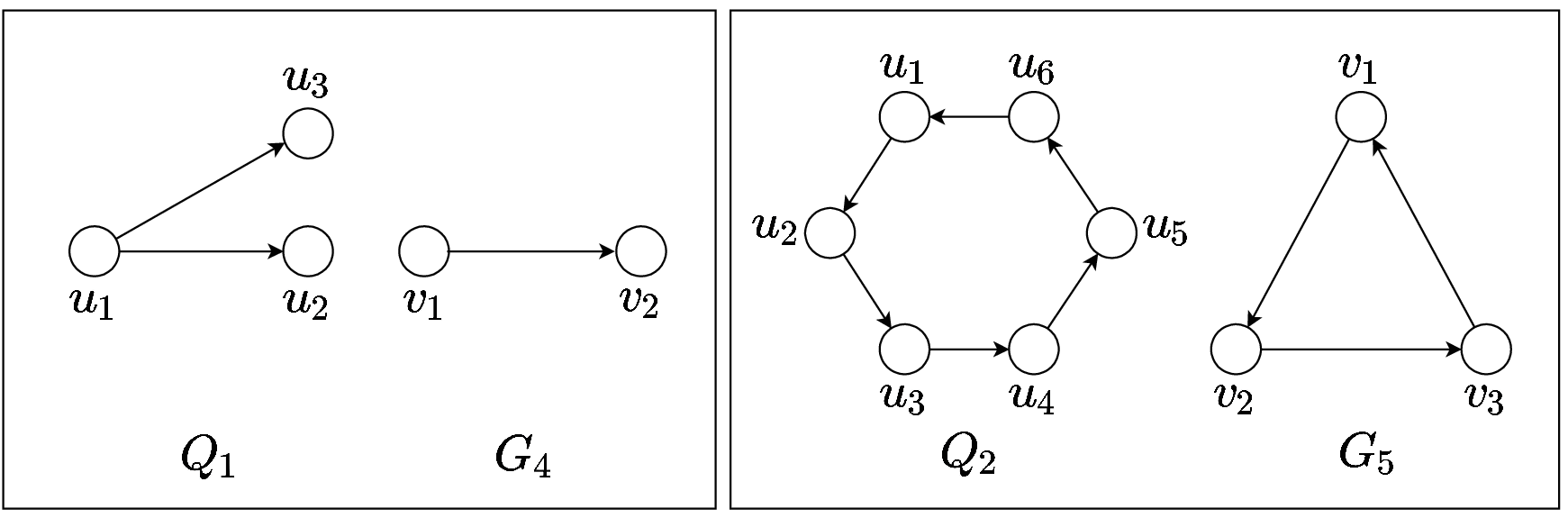}}
	\centering
	\vspace{-1.4ex}
	\caption{Cases for homomorphism} 
	\label{fig:difference-1}
	\vspace{-4ex}
\end{figure}

\stitle
(1) We need to ensure that two distinct vertices
in $Q$ can be mapped to the same vertex in $G$.
We illustrate two such scenarios in Figure~\ref{fig:difference-1}.

\sstab
(a) Consider pattern $Q_1$ and graph
$G_4$ 
in Figure~\ref{fig:difference-1}. 
Assume 
that all vertices and edges in $Q_1$ 
and   $G_4$ have the same label. 
There is a homomorphic mapping $\varphi_1$ from $Q_1$ to $G_4$
as follows: $\varphi_1(u_1){=}v_1$, $\varphi_1(u_2){=}h_1(u_3){=}v_2$.
Note that $u_2$ and $u_3$ are both mapped to  $v_2$ in $h_1$.
\looseness=-1

To handle such case with GNNs, we use {\em sets} to store received embedddings
of neighbors,
rather than multisets as in \IDGNN.
In this way, when embeddings of  $u_2$ and $u_3$ are the same,
	the duplicated one will removed, 
	and only embedding of $u_2$ or $u_3$ is used.
As a result, the embeddings of $u_1$ in $Q_1$ and
vertex $v_1$ in $G_1$ are the same,
from which \HGIN predicts that $Q_1$ is homomorphic to $G_4$. 
\looseness = -1 

\stab
(b)  
Consider $Q_2$ and $G_5$ in  Figure~\ref{fig:difference-1}, 
which are two cycles of length 6 and 3, respectively, 
and all vertices and edges carry the same label.
There exists a homomorphic mapping $\varphi_2$
from $Q_2$ to $G_5$ 
as follows:
$\varphi_2(u_1)=\varphi_2(u_4)=v_1$, 
$\varphi_2(u_2)=\varphi_2(u_5)=v_2$ and 
$\varphi_2(u_3)=\varphi_2(u_6)=v_3$.
However, \IDGNN
can only distinguish the two graphs, 
and cannot verify the existence 
of such a mapping, 
since $\kw{MSG}_1$ and $\kw{MSG}_0$
is not comparable.
To address this, we introduce the following 
two extensions. 
(a) We enforce that 
$\kw{MSG}_1$ is sufficiently larger than $\kw{MSG}_0$
when training \HGIN.
(b) 
We compute a set ${\mathcal L_{Q}}$ of 
lengths of cycles that contain
$u_1$ in pattern $Q$, and use $\kw{MSG}_1$ 
only when  
vertex $v_1$
also appears in a cycle of length  $l$
in $G$ satisfying that $l\in {\mathcal L}_{Q}$.
\looseness = -1

To show the effectiveness, consider the following two cases.

\sstab
(i) When pattern $Q$ contains a cycle of 
length  
$l$ involving $u$, whereas graph $G$ 
does not have such a cycle involving $v$, 
the computation of $u$'s embedding  in $Q$
invokes $\kw{MSG}_1$ more frequently.
Since  $\kw{MSG}_1$ is sufficiently larger than $\kw{MSG}_0$,
the embedding of $u$ is larger than that of 
$v$, indicating that no homomorphic mapping 
from $Q$ to $G$ exists.
\looseness = -1

\sstab
(ii) Conversely, if $G$ contains a cycle of length $l$
involving $v$, whereas $Q$ 
does not, 
$\kw{MSG}_0$ is invoked for vertices
within the cycle of length $l$ in
both $Q$ and $G$, which cannot result in that 
the embedding of $u$ is larger than 
that of $v$, thus retaining the semantics of homomorphic mappings.
Given $Q_2$ and $G_5$ in Fig.~\ref{fig:difference-1},
although $G_5$ contains a cycle of length 3  
involving $v_1$, whereas $Q_2$ 
does not have such a cycle involving $u_1$,
$\kw{MSG}_0$ is invoked to compute embeddings for both $Q_2$ and $G_5$.  
\looseness = -1

\sstab
(2) Since \IDGNN does not support edge labels, 
and cannot be directly used on the homomorphism problem,
we extend the R-GCN framework~\cite{r-GCN}.
The extension introduces a relation-specific 
transformation, to accommodate edge labels.
More specifically, we categorize edges 
by their labels, 
and assign different weights to different labels.
To address the directions of edges, 
we compute the embeddings
with incoming edges and outgoing edges
separately. 
Then we concatenate these embeddings to form the embedding of vertices.
\looseness=-1

\sstab
(3) \HGIN utilizes the order-embedding space 
to check the existence of homomorphic mappings~\cite{NeuroMatch},
involving comparison of values at 
each index of the embeddings.
However, the computed embeddings of vertices 
might include negative values.
To overcome this, we normalize the embeddings,
to ensure that all values are positive.
\looseness  =-1

\vspace{0.5ex}
Putting these together, we present \HGIN in 
Figure~\ref{fig:workflow}.

\etitle{Embeddings of $Q$}.
Given a vertex $u$, it first constructs
a subgraph $Q_u$  of $Q$, induced by
the $m$-hop neighbors of $u$,
where $m$ is a predefined parameter;
then it iteratively computes the embedding of 
each vertex $u_c$ in $Q_u$
as follows: (a) collects the embeddings
of $u_c$'s neighbors via a message passing function, 
and updates $u_c$'s embedding using the collected
embeddings via an aggregation function. More specifically, 
it computes the embedding
of $u_c$ as follows:
{‌\fontsize{8pt}{10pt}‌
	\begin{align}
		h^0_{u_c} &= e^f_x\\
		h^k_{u_c, i} &= \kw{AGG}^k(\sum\nolimits_{r\in R_G}\kw{MSG}^{k,r}_{{\bf 1}[u_n=u]}(\{h^{k-1}_{u}, u_n{\in} N^-_r(u_c)\}),
		h^{k-1}_{u_c})\\
		h^k_{u_c,o} &= \kw{AGG}^k(\sum\nolimits_{r\in R_G}\kw{MSG}^{k,r}_{{\bf 1}[u_n=u]}(\{h^{k-1}_{u}, u_n{\in} N^+_r(u_c)\}),
		h^{k-1}_{u_c})\\
		h^k_{u_c} &= \kw{COM}(h^k_{u_c,i},h^k_{u_c,o}).
	\end{align}
}
Here, (1) $N_r^-(u_c)$ (resp. $N_r^+(u_c)$)
is the set of 
neighbors of $u_c$
which have an 
outgoing edge to $u_c$ 
(resp. an incoming edge from $u_c$) labeled $r$;
and (2) the message passing function
$\kw{MSG}^{k,r}_{{\bf 1}[u_n=u]}$ collects and transforms 
embeddings of neighbors of $u_c$;
in this paper, we adopt 
matrix multiplication as the message passing functions;
more specifically, 
$\kw{MSG}^{k,r}_{{\bf 1}[u_n=u]}(N^-_r)$ transforms the message to 
\begin{align}\sum\nolimits_{u_n\in N^-_r}W^{k,r}_{{\bf 1}[u_n=u]}h^{k-1}_{u_n};
\end{align}
here $N^-_r$ is the set of outgoing neighbors;
the message passing functions for incoming neighbors are defined similarly;
(3) ${\bf 1}[u_n=u]$ is an indicator function that
returns $1 $ if $u_n$ and $u$ are the same vertex,
and 0 otherwise; 
that is, if the neighbor $u_n$ is the center vertex
$u$ of $Q_u$, then the message
passing function is $\kw{MSG}^{k,r}_1$;
otherwise, the message passing function is  $\kw{MSG}^{k,r}_0$;
(4) $\kw{AGG}^k$ combines the embeddings
of neighbors and the current embedding of $u_c$ as follows:
\looseness = -1
\begin{align}\kw{AGG}^k(\kw{MSG}, e_2) = W_1e_2 + \kw{MSG};
\end{align}
and (5) $\kw{COM}$ updates the embedding of 
$u_c$ by concatenating the 
embeddings of its incoming neighbors 
and outgoing neighbors.
\looseness = -1

\etitle{Embeddings of $G$}.
For a pattern $Q$ and a graph $G$,  given vertices $u$ in $Q$ and $v$ in $G$,
it first compute the sets ${\mathcal L}_{Q}^u$ and ${\mathcal L}_{G}^v$ 
of all cycles of length $l$ in $Q$ and $G$ containing $u$ and $v$, respectively,
where $l$ is bounded by $m$, \ie the number of layers in \HGIN.
Let $G_v$ be the subgraph of $G$, induced by the $m$-hop neighbors of $v$.
Then it iteratively computes the embedding of each vertex $v_c$ in $G_v$ 
in the same way as it does for  embedding $Q$, except that 
the message-passing functions
for each neighbor $v_n$ of vertex $v_c$ in $G_v$ 
is $\kw{MSG}^{k,r}_{{\bf 1}[v_n=v][k\in {\mathcal L}_{Q}^u]}$.
Here ${\bf 1}[v_n{=}v][k{\in} {\mathcal L}_Q^u]$ 
is a Boolean function,
which returns $1$ when 
(1) the neighbor $v_n$ is the center 
$v$ of $G_v$,
and (2) $Q$ has a cycle of length $k$ 
containing the center $u$ in $Q_u$, and $0$
otherwise.
\looseness=-1

Note that when $Q_u$ does not
have a cycle of length $k$ 
that contains the center $u$,
the massage-passing function is 
$\kw{MSG}^{k,r}_0$, regardless of whether 
$v_n$ is the center $v$ of $G_v$ or not.

\etitle{Normalization}.
After \HGIN generates embeddings of vertices, 
it conducts L2 normalization
to ensure that all embeddings 
are positive.
More specifically, 
for each vertex $v$,
let $e_v=(x_1, \ldots, x_d)$ be its embedding generated via \HGIN,
and its L2 norm is  $\cdd{e_v}_2=\sqrt{x^2_1+\ldots+x^2_d}$.
Then the normalized embedding $e^N_v$ of
the original embedding $e_v$
is defined as 
$e^N_v:= \frac{|e_v|}{\cdd{e_v}_2}$.

\etitle{Prediction Function}.
Given the embedding $e_u{=}(a_1, \ldots, a_d)$
of vertex $u$ in $Q$ and the embedding $e_v=(b_1, \ldots, b_d)$
of $v$ in $G$, 
it exploits the order embedding space~\cite{Order-embedding-1,NeuroMatch},
to check the existence of homomorphic mapping 
from $Q$ to $G$ as follows:
$Q$ is homomorphic to $G$
if and only if 
$a_i\leq b_i$ for all $i\in [1, d]$.
To improve the generalization of \HGIN, 
a threshold $t$ is used to
bound the distance between $e_u$
and $e_v$. 
More specifically, the prediction function $f$ is defined as:
\begin{equation}\label{eq8}
	f\left(e_u, e_v\right)= \begin{cases}1 & \text{ if } \mathcal{M}\left(e_u, e_v\right)\leq t; \\ 0 & \text { otherwise }.\end{cases}
\end{equation}
Here, function $\mathcal{M}(e_u, e_v) $ 
is defined as $ \cdd{\max(0, e_u-e_v)}_2^2$.

\etitle{Loss function}.
To train \HGIN, 
we enhance the max-margin loss function
by integrating
message-passing functions.
This enforces vertex identities 
in \IDGNN,
by ensuring that  
$M^{k,r}_1$ is sufficiently larger than 
$M^{k,r}_0$.
When $Q$ has a cycle of length 
$l$ containing $u$, but $G$ does not 
have such a cycle containing $v$, 
computing the embedding of $u$ 
invokes message-passing functions $M^{k,r}_1$
more frequently than 
computing the embedding of $v$.
Thus, the embedding of $u$ is larger than 
that of $v$, 
implying that $Q$ is not homomorphic to $G$.
More specifically, the max-margin loss function is defined as follows.
\begin{align}\label{eq6}
	\mathcal{L} &= \sum\nolimits_{(Q, u, G, v) \in P}(\mathcal{M}(h_u, h_v) - \alpha) \nonumber\\
	&\hspace{2ex} +\sum\nolimits_{(Q, u, G, v) \in N}\text{max}\{0, \alpha - \mathcal{M}(h_u, h_v)\} \nonumber\\
	&\hspace{2ex} +\sum\nolimits_{k\leq m}\sum\nolimits_{r\in R} \frac{1}{\kw{MSG}^{k,r}_1-\kw{MSG}^{k,r}_0}, 
\end{align}
where (1) $P$ is the set of 
positive examples
$(Q, u, G, v)$ such that 
there exists a homomorphic mapping $\varphi$
from $Q$ to $G$ such that
$\varphi(u)=v$, (2) $N$ is the set of negative examples $(Q, u, G, v)$ 
such that there {\em does not} exist a homomorphic mapping $\varphi$
from $Q$ to $G$ such that
$\varphi(u){=}v$, 
and (3) $\alpha$ is the margin threshold 
for the prediction function.
\looseness = -1

\eetitle{Remark}.
The loss function $\mathcal{L}$ 
differs from the loss function 
for subgraph matching in~\cite{NeuroMatch}
as follows.

\stab
(1) We add the term $\frac{1}{\kw{MSG}^{k,r}_1-\kw{MSG}^{k,r}_0}$
to ensure that
$\kw{MSG}^{k,r}_1$ 
is sufficiently larger than $\kw{MSG}^{k,r}_0$ for each edge label $r$.
Intuitively, sufficiently 
large  $\kw{MSG}^{k,r}_1$  allocates more weights
to  the appearance of center vertices.
Combining with the set ${\mathcal C}$ of cycle sizes,
\HFrame can distinguish more negative pairs
than existing solutions,
\ie~\HFrame has more expressive power
(see Section~\ref{sec-properties}).
\looseness = -1

\stab
(2) We add the term $-\alpha$ to penalize
errors incurred by positive training data.
This guarantees that the loss function 
is 1-Lipschitz, which is used to prove the generalization bound
for \HFrame (Theorem~\ref{thm-generalization-bound}).
\looseness=-1

	\stitle{Complexity}. 
	HFrame takes $O(|Q||G|+m\delta_{max}(|Q|+|G|)(f(|Q|)+f(|G|)))$ time 
	and $O(m\delta_{max}d(|G|+|Q|)f_M)$ space for inference, 
	and $O((|P|{+}|N|)(|Q||G|{+}m\delta_{max}(|Q|{+}|G|)(f(|Q|)+f(|G|))))$ time 
	and $O((|P|{+}|N|)m\delta_{max}d(|G|{+}|Q|)f_M)$ space for training.
	Here (1) $m$ is number of layers, 
	(2) $\delta_{max}$ is the maximum degree of $Q$ and $G$, 
	(3) $f$ is a lower-degree polynomial
	denoting embedding time, 
	(4) $d$ is dimension of embeddings,
	(4) $f_M$ is the size of \HGIN, and 
	(5) $|P|$ (resp. $|N|$) is the number of positive 
	(resp. negative) training data.
	\looseness = -1

\section{Theoretical analysis}
\label{sec-properties}
We first study the expressive power
of \HFrame and then establish a generalization error bound for 
it.
Since \HFrame consists of 
\dual and \HGIN, 
we start with comparing their expressive power,
to justify their combination. 
See Appendix for detailed proofs.
\looseness  = -1

\stitle{(1) Expressive power}.
We consider the following three cases.

\etitle{(a) \dual vs. vanilla \HGIN}.
We first show that \dual provides 
an upper bound for the expressive power 
of vanilla \HGIN  ($\HGIN^\kw{v}$),
which uses a unified message passing function for all vertices. 
More specifically, 
the same message passing function $\kw{MSG}^{k,r}$
is applied to all neighbors
of vertices. Intuitively, both $\HGIN^\kw{v}$ and \dual
inspect only neighbors of a vertex,
and have similar expressive power.
We prove that there exists
a $\HGIN^\kw{v}$ that is as expressive as \dual. 
\looseness = -1

\begin{theorem}
	Given a pattern $Q$, a graph $G$,
	a vertex $u$ in $Q$ and a vertex $v$ in $G$, 
	if $\HGIN^\kw{v}$
	returns \false,
	then $(u, v)\not\in S_{(Q, G)}$, 
	\ie~\dual also determines that
	there  exists no homomorphism mapping $\varphi$
	from $Q$ to $G$ such that $\varphi(u)=v$,
	even when all functions $\kw{MSG}^{k,r}$
	and $\kw{AGG}^k$
	$(k\leq m)$ in $\HGIN^\kw{v}$
	are injective and monotonic.
	\label{thm-upper-bound}
\end{theorem}
\proof
We prove it by contradiction.
Assume by contradiction 
that there  exist a pattern $Q$, a graph $G$,
a vertex $u$ in $Q$ and a vertex $v$ in $G$, 
such that $\HGIN^\kw{v}$ returns \false, 
but \dual returns \true, \ie $(u, v)\in S_{(Q, G)}$. 
We deduce a contradiction as follows.
(a) Computing
embeddings of $u$ and $v$ can be represented by
trees of height $m$, where $m$ is
the number of layers in $\HGIN^\kw{v}$.
(b) Because $\kw{MSG}^{k,r}$ and $\kw{AGG}^k$
$(k\leq m)$ in $\HGIN^v$
are injective and monotonic,
there exists a path $p_0$ in the
computation tree for $u$ such that 
no path from $v$ carries the same labels as $p_0$.
However, if these hold, \dual can also identify 
such path and return \false, a contradiction.
\looseness=-1
\eop

\etitle{(b) \dual vs. \HGIN}.
When \HGIN uses different message passing functions
for vertices, 
it can distinguish the center from  other vertices
and 
has a higher expressive power than \dual.
\looseness=-1

\begin{theorem}
	\HGIN is more expressive than \dual.
	\label{thm-hgin-dual}
\end{theorem}

\proof
(1) We first show that given a pattern $Q$, a graph $G$,
a vertex $u$ in $Q$ and a vertex $v$ in $G$,
when \dual returns \false, \ie $(u, v)\not\in S_{(Q, G)}$,
there is a $\HGIN$ that returns \false
given the input.
\looseness = -1

\stab
(2) We next show that $\HGIN$ can distinguish 
more graphs than \dual.
Consider graphs $G_2$ and $G_3$  
in Figure~\ref{fig:difference},
extended from~\cite{you2021identity}.
(a) There is no homomorphic mapping 
from $G_3$ to $G_2$, since $G_3$ contains
triangles while  $G_2$ does not.
(b) \HGIN returns \false given $w_1$
and $v_1$, since 
$w_1$ is 
in a cycle of length 3
while $v_1$ does not, which results in a larger embedding 
of  $w_1$ 
than that of $v_1$.
(c) \dual confirms that $(w_1, v_1){\in} S_{(G_3,G_2) }$,
since $G_2$ and $G_3$ are regular, all vertices have similar neighborhoods,
and $S_{(G_3, G_2)}=V_{G_3}{\times} V_{G_2}$.
\looseness=-1
\eop

\etitle{(c) \dual vs. \HGIN of finite dimensions}.
\HGIN represents vertices
with embeddings 
of finite dimensions.
For graphs with a large number of labels
 ({\em e.g.}, more edge labels than the embedding dimension), 
we can  construct a pattern 
$Q$  and a graph $G$ with different label sets
such that 
(1) the embedding of $Q$ is smaller than that of $G$, 
but (2) $Q$ is not homomorphic to $G$.
However, \dual can distingush graphs with any number of labels, by directly comparing their labels.

\begin{theorem}
	When the dimension of embeddings is finite, 
	there exist two graphs that \dual
	can distinguish, but \HGIN cannot.
	\label{thm-more-expressive-1}
\end{theorem}

\proof
Assume that \HGIN generates $d$-dimensional embeddings. 
(a) We select $2d$ distinct labels
from the label set $\Gamma$
and compute their embeddings using \HGIN; then
we choose $d$ embeddings $e_1, \ldots, e_d$
 such that each $e_i$ has the maximum $i$-th element.
(b) We construct a star pattern $Q_s$ centered at  $v$, with $d$ edges,
each labeled with one of the $d$ picked labels.
We then deduce a contradiction by comparing
the embeddings of $Q_s$ with that of the $2d$ distinct labels.
\eop

Theorems~\ref{thm-hgin-dual} and~\ref{thm-more-expressive-1} imply that 
\HGIN and \dual  have incomparable expressive power with finite embedding dimensions.
We can improve pattern matching by combining \dual with \HGIN.

\stitle{(2) Generalization bound}.
We give a generalization error bound
for \HFrame to measure its prediction accuracy on unseen data.
\looseness=-1

\vspace{0.1ex}
\begin{theorem}
	The empirical Rademacher complexity of \HFrame
	is $\hat{\mathcal{R}}(\mathcal{T})\leq\frac{8}{N_T}+
	\frac{12}{\sqrt{N_T}}\sqrt{2|R|d^2\log T}$, where
	$\mathcal{T}$
maps the training data to its loss,
	and $T=16M\sqrt{{N_T}d}\max\{B_xB_1,{\mathcal R}B_2\}$.
	\label{thm-generalization-bound}
\end{theorem}
\vspace{0.1ex}

Here, (1) $|R|$ is the number of edge labels in $Q$ and $G$;
(2) $N_T$ is the  
size of training data;
and (3) $M$ and ${\mathcal R}$ are numbers
determined by functions in \HGIN.
Assume that the norms of $W_1$, $W^{k,r}_i$ and $h^k_{u_c} $
are bounded by $B_1$, $B_r$ and $B_x$, respectively.

\eetitle{Remark}.
We extend 
the proof of
the Rademacher complexity for GNNs in~\cite{Gen-bound}.
(1) \HFrame takes a pattern $Q$ and a graph $G$
as input, while GNNs in~\cite{Gen-bound} 
process only one graph;
(2) \HFrame handles patterns and graphs with both labels and directions, which are not explored 
in~\cite{Gen-bound};
(3) \HGIN uses distinct message-passing functions 
for different neighbors,
a feature absent in GNNs of~\cite{Gen-bound};
(4) \HGIN uses the max-margin loss function
to compare embeddings of vertices, which
complicates the analysis of \HGIN.
\looseness = -1

\section{Experimental study}
\label{exp:sec-exp}
Using real-life and synthetic graphs, 
we evaluated the (1) effectiveness, (2) efficiency,  
(3) parameter sensitiveness and (4) generalization 
	of  the proposed  framework 
	\HFrame.
	(5) We also conducted an ablation study on \HFrame,
and (6) applied \HFrame in 
	subgraph homomorphism  enumeration and frequent pattern mining.
\looseness=-1

\stitle{Experimental settings}.
We start with the settings.

\etitle{Data Graphs.} We used six real-life graphs:
(1) Citation, a citation network with 1,397,240 vertices and 3,016,539 edges,
carrying 16,415 labels~\cite{Citation};
(2) Citeseerx, another citation network with 6,540,401 vertices and 15,011,260 edges,
carrying 984 labels~\cite{Citeseerx};
(3) DBLP, a collaboration network with 205,783 vertices and 301,561 edges,
carrying 11 labels~\cite{DBLP};
(4) IMDB, a movie database modeled as graph, with 5,101,080 vertices and 5,174,035 edges,
carrying 38 labels~\cite{IMDB};
(5) DBpedia, a knowledge graph with 5,174,170 vertices and 17,494,747 edges
with 9,591 labels~\cite{DBpedia}; and
(6) YAGO, a knowledge graph with 3,489,226 vertices and 7,351,194 edges
carrying 15 labels~\cite{YAGO}.

For synthetic dataset, we 
generated a graph 
$G=(V, E, L)$, controlled by
the number of vertices $|{V}|$ (up to 20M),
the number of edges $|{E}|$ (up to 40M)
and the number of labels $|{L}|$ (up to 500).

\etitle{Baselines}. 
We categorized the baselines as follows. 

\eetitle{Algorithms}.
We compared \HFrame with the following algorithms
for subgraph matching: 
(1) the edge-join approach (\kw{EJ})~\cite{cheng2010graph}, 
which computes the occurrences for each edge of 
the query in the data graph, 
 finds an optimized plan to join these binary relations, 
and finally evaluates the query following the plan; 
(2) the path-join approach (\kw{PJ})~\cite{sun2019scaling}, 
which first generates solutions to each root-to-leaf path of the given query, 
and then produces the answer by 
joining  solutions of these paths;
(3) \kw{SIM}-\kw{TD}~\cite{wu2023novel}, 
which encodes all possible homomorphisms from a pattern to the graph with the concept of answer graph;
(4) a pattern matching algorithm 
\kw{DualSim}~\cite{ma2011capturing} (see Section~\ref{sec-pre});
and (5) popular graph DBMS~\kw{Neo4j}~\cite{Neo4j}.

Among these algorithms, \kw{EJ}, \kw{PJ}, \kw{SIM}-\kw{TD} 
and \kw{Neo4j} are exact algorithms, 
and \kw{DualSim} is an approximate algorithm.
For a fair comparison, we modified 
	\kw{EJ}, \kw{PJ}, \kw{SIM}-\kw{TD} 
	and \kw{Neo4j} to return only one 
	match, 
	as we focus on decision problem
	rather than enumeration.
	\looseness  =-1

\vspace{-0.3ex}
\sloppy
\eetitle{ML models}. We compared with the following 
ML models: (6) \kw{NeuroMatch}~\cite{NeuroMatch}, which applies 
GNN to compute the embedding, and exploits
the order embedding space to make predictions
for subgraph isomorphism;
(7) \kw{D2Match}~\cite{Sub-ML-1},
which proposes a degeneracy procedure
that frames subgraph matching as a subtree matching problem;
and (8) \kw{DMPNN}~\cite{DMPNN}, 
which develops dual message passing neural networks 
to improve substructure representation.

\vspace{-0.3ex}
\etitle{Training data}.
We constructed training data by sampling from large data graphs
following~\cite{Sub-ML-1, NeuroMatch}.
It consists of both positive and negative data.
(1) We constructed a set $P$ of positive data $(Q_p, u, G_p, v)$, 
\ie there exists a homomorphic mapping $\varphi$
from $Q_p$ to $G_p$ such that $\varphi(u)=v$.
For each graph $G$, 
we (a) first  randomly extract a subgraph $G_p$ 
	by performing  BFS  from a random vertex on $G$
	in 5 to 10 steps,
 (b)  select a connected graph $Q_p$ from $G_p$ as the pattern, 
where the number of nodes in $Q_p$ is between 3 and 20,
(c) randomly copy vertices and edges in $Q_p$ 
to enforce the homomorphic semantics,
and (d) pick one vertex $u$ from $Q_p$ as the pivot and 
set the same vertex in $G_p$ as the pivot $v$.
(2) We also construct a set $N$ of 
{\em negative data} $(Q_n, u, G_n, v)$, 
such that no homomorphic mapping exists from $u$
to $v$.
To this end, we first extract  positive data
$(Q_p, u, G_p,v)$ as above, 
and then calibrate the pattern $Q_p$
to create  pattern $Q_n$ that has no match  
in $G_p$, by adding edges randomly to $Q_p$ or perturbing the pivot in $G_p$.

To make ensure that \HFrame is robust, we set the ratio of positive to 
negative instances as $1:3$.
Moreover, we employed \kw{Neo4j} to 
compute the ground-truth between the patterns 
and the graphs.
\looseness=-1

We sampled 20,000 pairs of query and subgraph from each graph, 
\ie 5,000 positive and 15,000 negative examples, 
and split each dataset into training, validation 
and test sets at a ratio of 8:1:1.
As \kw{EJ}, \kw{PJ} and \kw{D2Match} do not 
support edge labels, 
when comparing with these baselines, 
we removed all edge labels 
from patterns  and graphs for fair comparison.
It should be noted that, since \kw{SIM}-\kw{TD} is only applicable to tree structure, 
every query was decomposed into multiple tree queries before conducting the experiments on \kw{SIM}-\kw{TD}.

\vspace{-0.5ex}
\etitle{Evaluation Metric}.
We tested accuracy of 
\HFrame
with the metric:
$ \text{Acc} = (TP + TN)/(TP + TN + FP + FN)$,
where $TP$, $TN$, $FP$ and $FN$ denote
the numbers of true positives, true negatives, false positives and false negatives, respectively.
In terms of efficiency, we measure the end-to-end CPU time 
	for exact algorithms (\ie~\kw{EJ}, \kw{PJ}, \kw{SIM}-\kw{TD} and \kw{Neo4j)};
for \HFrame, we measure CPU time for \dual and GPU time for \HGIN.
When there exists no matching for  a query, the exact algorithms will exhaustively traverse 
the entire search space before returning false. 
To ensure the fairness on efficiency comparison, 
we  imposed a 30-second time limit on the runtime. 
\looseness=-1

\vspace{-0.5ex}
\etitle{Experiment Environments.} 
We run the experiments on a server running 
Ubuntu 11.4.0 with Intel Xeon E5-2680v4 CPU, 
NVIDIA Tesla V100 PCIe 32GB GPU and 512GB RAM.
\HGIN  training and inference  are performed on GPU, 
while the remaining components of \HFrame run  on CPU. 
By default, we set the number of layers $m=5$, 
the dimension of embeddings $d = 64$, the margin $\alpha  = 1.5$ and the threshold $t = 0.1$.
We set the number of iterations for \dual to 2.
We ran each test 5 times, and report the average here.

\stitle{Experimental findings}. We next report our findings.

\stitle{Exp-1: Accuracy}.
We first evaluated the accuracy of \HFrame.
Since \kw{EJ}, \kw{PJ}, \kw{SIM}-\kw{TD}
and \kw{Neo4j} are exact algorithms, 
their accuracy are 1.
In the following, we only compared \HFrame
with \dual, 
\kw{NeuroMatch}, \kw{D2Match} and \kw{DMPNN}. 
As shown in Table~\ref{tab:accuracyresults},
(1) \HFrame achieves the best performance on all graphs.
On average, 
\HFrame improves 
\dual, \kw{NeuroMatch}, \kw{D2Match}
and \kw{DMPNN} by 54.67\%, 18.90\%, 17.02\% and 
16.88\%, respectively.
This is because \HFrame (a) invokes \dual
to filter candidates, (b) adopts sets
to aggregate messages,
and (c) uses different message passing functions 
for different vertices.
Model \kw{D2Match} runs out of memory on Citeseer, DBpedia and YAGO, marked as OOM.
(2) \HFrame is practical. 
The accuracy of \HFrame
can achieve 0.921 on all graphs, while the accuracy 
of all ML-based solutions and \dual
are lower than 0.9.
This is because \HFrame is more expressive
than existing solutions (see Section~\ref{sec-properties}). 

\begin{table}[t]
	\caption{Accuracy Comparison Results (Acc)}
	\label{tab:accuracyresults}
	\vspace{-1.5ex}
	\centering
	\resizebox{1.0\columnwidth}{!}{
		\begin{tabular}{c|ccccccc} \hline
						\bf{Method}
			&Citation&Citeseerx&DBLP&IMDB&DBpedia&YAGO&Synthetic \\ \hline
			\dual     			& 0.669 & 0.573 & 0.674 & 0.590 & 0.601 & 0.575 & 0.692  \\ 
			\kw{NeuroMatch}     & 0.804 & 0.803 & 0.837 & 0.798 & 0.782 & 0.783 & 0.857  \\ 
			\kw{D2Match}        & 0.804 & OOM   & 0.833 & 0.784 & OOM   & OOM   & 0.881  \\ 
			\kw{DMPNN}          & 0.815 & 0.812 & 0.842 & 0.806 & 0.805 & 0.794 & 0.889  \\ 
			\HFrame             & {\bf 0.954} & {\bf 0.982} & {\bf 0.969} & {\bf 0.963} & {\bf 0.969} & {\bf 0.921} & {\bf 0.974}  \\ \hline
			$\HFrame_{\kw{WS}}$ & 0.895 & 0.952 & 0.965 & 0.961 & 0.941 & 0.902 & 0.939  \\ 
			$\HFrame_{\kw{MS}}$ & 0.837 & 0.826 & 0.838 & 0.812 & 0.805 & 0.788 & 0.883 \\ 
			$\HFrame_{\kw{WD}}$ & 0.888 & 0.950 & 0.926 & 0.919 & 0.866 & 0.855 & 0.929  \\ 
			$\HFrame_{\kw{WN}}$ & 0.938 & 0.963 & 0.952 & 0.950 & 0.918 & 0.893 & 0.951 \\ 
			$\HFrame_{\kw{WC}}$ & 0.930 & 0.920 & 0.945 & 0.937 & 0.921 & 0.884 & 0.942 \\
			$\HFrame_{\kw{WG}}$ & 0.943 & 0.950 & 0.957 & 0.946 & 0.944 & 0.899 & 0.965 \\ \hline
		\end{tabular}
	}
\end{table}

\stitle{Exp-2: Efficiency}.
As shown in Table~\ref{tab:efficiency},
(1) \HFrame is faster than exact algorithms.
On average, \HFrame outperforms \kw{SIM}-\kw{TD}, \kw{EJ} and \kw{PJ}
by 28.36$\times$, 101.91$\times$ and 85.06$\times$, respectively.
(2) \HFrame is slightly slower than
\kw{NeuroMatch}
since \HFrame calls \dual to filter candidates, 
and inspects cycles.
(3) Although \kw{NeuroMatch} 
is the fastest, 
its
accuracy is much lower than \HFrame.
Hence, \HFrame strikes a balance between
expressive power and efficiency.

\begin{table}[t]
	\caption{Efficiency Comparison results (Time in s/query)} 
\label{tab:efficiency}
\vspace{-1.5ex}
\centering
\resizebox{1.01\columnwidth}{!}{
	\begin{tabular}{c|ccccccc} 
			\hline
				\bf{Method}
		&Citation&Citeseerx&DBLP&IMDB&DBpedia&YAGO&Synthetic\\\hline
		\kw{EJ}              & 1.711 & 2.286 & 2.017 & 2.111 & 1.929 & 2.221 & 1.516 \\ 
		\kw{PJ}              & 1.545 & 1.856 & 1.803 & 1.746 & 1.735 & 2.219 & 1.101 \\ 
		\kw{SIM}-\kw{TD}     & 0.451 & 0.716 & 0.659 & 0.582 & 0.511 & 0.674 & 0.325 \\ 
		\kw{Neo4j}           & 0.576 & 0.885 & 0.617 & 0.594 & 0.686 & 0.639 & 0.513 \\ 
		\kw{DualSim}         & 0.056 & 0.083 & 0.054 & 0.069 & 0.059 & 0.095 & 0.051 \\ 
		\kw{NeuroMatch}      & {\bf 0.005} & {\bf 0.029} & {\bf 0.007} & {\bf 0.006} & {\bf 0.012} & {\bf 0.022} & {\bf 0.017} \\ 
		\kw{D2Match}         & 0.039 & OOM   & 0.045 & 0.038 & OOM   & OOM   & 0.040 \\ 
		\kw{DMPNN}           & 0.016 & 0.058 & 0.015 & 0.011 & 0.026 & 0.051 & 0.035 \\ 
		\HFrame              & 0.012 & 0.048 & 0.013 & 0.014 & 0.019 & 0.039 & 0.034 \\ \hline
		$\HFrame_{\kw{WS}}$  & 0.005 & 0.030 & 0.007 & 0.006 & 0.012 & 0.028 & 0.024 \\ 
		$\HFrame_{\kw{MS}}$  & 0.012 & 0.046 & 0.013 & 0.014 & 0.019 & 0.039 & 0.023 \\ 
		$\HFrame_{\kw{WD}}$  & 0.012 & 0.044 & 0.012 & 0.013 & 0.019 & 0.039 & 0.024 \\ 
		$\HFrame_{\kw{WN}}$  & 0.013 & 0.048 & 0.013 & 0.014 & 0.019 & 0.040 & 0.024 \\ 
		$\HFrame_{\kw{WC}}$  & 0.012 & 0.043 & 0.012 & 0.013 & 0.018 & 0.037 & 0.020 \\ 
		$\HFrame_{\kw{WG}}$  & 0.013 & 0.048 & 0.013 & 0.014 & 0.019 & 0.040 & 0.024 \\ \hline
	\end{tabular}
}
\end{table}

For a fair comparison, we evaluate \HFrame and 
exact algorihms  \kw{SIM}-\kw{TD}, \kw{EJ}, \kw{PJ} only  on positive instances. 
	As shown in Table~\ref{tab:efficiency-pos},  
\HFrame consistently outperforms exact algorithms in efficiency.
	We further report the average training, inference and 
	\dual time for each query on CPU in Table~\ref{tab:efficiency-cpu-2}. 
The average matching time 
(\ie times for inference and \dual) 
 of \HFrame over all datasets is 0.074s on CPU and 0.026s on GPU,
 indicating no order-of-magnitude difference.  
 
\begin{table}[t]
	\caption{Efficiency  on positive examples (Time in s/query)}
	\label{tab:efficiency-pos}
	\vspace{-1.5ex}
	\centering
	\resizebox{1.01\columnwidth}{!}{
		\begin{tabular}{c|ccccccc} \hline
			\bf{Method}
			&Citation&Citeseerx&DBLP&IMDB&DBpedia&YAGO&Synthetic\\\hline
			\kw{EJ}              & 1.642 & 2.197 & 1.926 & 2.002 & 1.840 & 2.144 & 1.406 \\ 
			\kw{PJ}              & 1.489 & 1.765 & 1.743 & 1.671 & 1.638 & 2.118 & 0.997 \\ 
			\kw{SIM}-\kw{TD}     & 0.428 & 0.683 & 0.591 & 0.514 & 0.476 & 0.619 & 0.302 \\ 
			\HFrame              & 0.013 & 0.048 & 0.013 & 0.015 & 0.019 & 0.040 & 0.034 \\ \hline
		\end{tabular}
	}
\end{table}

\begin{table}[t]
	\caption{Efficiency of \HFrame on CPU}
	\label{tab:efficiency-cpu-2}
	\vspace{-1.5ex}
	\centering
	\resizebox{1.01\columnwidth}{!}{
		\begin{tabular}{c|ccc} \hline
			\bf{Dataset} & Avg Training (s/epoch) & Avg Inference (s) & Avg DualSim (s) \\ 
			\hline
			Citation & 0.028 & 0.019 & 0.007 \\
			Citeseerx & 0.207 & 0.138 & 0.018 \\
			DBLP & 0.051 & 0.030 & 0.006 \\
			IMDB & 0.044 & 0.025 & 0.008 \\
			DBpedia & 0.093 & 0.051 & 0.007 \\
			YAGO & 0.199 & 0.106 & 0.011 \\
			Synthetic & 0.119 & 0.079 & 0.010 \\
			\hline
		\end{tabular}
	}
		\vspace{-2ex}
\end{table}

\begin{table*}[tb!]
	\caption{Parameter sensitive of \HFrame}
	\label{tab:results-sensitive-0}
	\centering
	\resizebox{1.5\columnwidth}{!}{
		\begin{tabular}{ccc|ccc|ccc|ccc|ccc}\hline
			$|Q|$ & Acc & T(ms)& $m$ & Acc &T(ms) & $t$ & Acc & T(ms) & $d$ & Acc & T(ms) & $\delta_{\rm avg}$ & Acc & T(ms)\\ \hline
			10 & {\bf 0.979} & {\bf 11.96} & 2 & 0.827 & {\bf 9.92} & 0.01 & 0.883 & 12.42& 16 & 0.919 & {\bf 8.10} & $[1, 2)$ & {\bf 0.978} & {\bf 30.06} \\ 
			15 & 0.970 & 12.25 & 3 & 0.881 & 10.53 & 0.05 & 0.952 &  12.42 & 32 & 0.930 & 10.59 & $[2, 3)$ & 0.974 & 35.11 \\ 
			20 & 0.946 & 12.47 & 4 & 0.925 & 11.31 & 0.1  & {\bf 0.954} &  12.42& 64 & {\bf 0.954} & 12.42 & $[3, 4)$ & 0.963 & 41.69 \\ 
			25 & 0.917 & 12.70 & 5 & {\bf 0.954} & 12.42 & 0.2  & 0.901 & 12.42& 128 & 0.903 & 20.87 & $[4, 5)$ & 0.962 & 55.13 \\ 
			30 & 0.902 & 13.12 & 6 & 0.929 & 13.59 & 0.3  & 0.908 & 12.43 & 256 & 0.892 & 28.77 & $[5, 6)$ & 0.941 & 72.20 \\ 
			35 & 0.884 & 13.51 & 7 & 0.903 & 14.62 & 0.5  & 0.878 &  12.42 & 512 & 0.758 & 41.26 & $[6, +\infty)$ & 0.864 & 93.85\\ 
			\hline
		\end{tabular}
	}
\end{table*}

\stitle{Exp-3: Parameter sensitiveness}.
We next evaluated the impacts of the sizes $|Q|$ of 
patterns (\ie $|V_Q|+|E_Q|$), the number $m$ of layers 
in \HGIN, the threshold $t$, the dimension $d$
of embeddings, 
and the average degree $\delta_{\rm avg}$ of graphs on the performance of \HFrame.
 We conducted experiments on dataset Citation.
\looseness = -1

\etitle{Varying $|Q|$}. 
We varied $|Q|$ from 10 to 35.
As shown in Table~\ref{tab:results-sensitive-0},
(1) when $|Q|$ increases,
the accuracy of $|Q|$  decreases, as expected, 
since making prediction is more challenging
for large patterns;
and (2) \HFrame becomes slower when $|Q|$ increases, 
as expected,
since it takes more time to compute embeddings.

\etitle{Varying $m$}.
We varied $m$ from 2 to 7.
As shown in Table~\ref{tab:results-sensitive-0},
(1) when 
$m$ increases,
\HFrame gets slower, 
as expected, since it takes more time to compute embeddings;
and (2) the accuracy of \HFrame first increases
and then decreases, \ie its optimal performance achieves
when $m=5$.
This is because when $m$ is small, 
\HFrame does not have sufficient information 
to make a prediction;
while when $m$ is large, 
\HFrame cannot distinguish neighbors
distant from the pivot.
\looseness = -1

\etitle{Varying $t$}.
We varied the threshold $t$ from 0.01 to 0.5.
As shown in Table~\ref{tab:results-sensitive-0},
(1) the runtime of \HFrame is insensitive
to $t$, 
since the threshold is used only when 
the embeddings of vertices have been computed;
and (2) the accuracy of \HFrame first increases
and then decreases, and its optimal performance achieves
when $t=0.1$.
This is because (a) when $t$ is small, 
\HFrame returns \true only when 
it correctly makes predictions, 
and the recall of \HFrame is low;
and (b) when $t$ is large, 
\HFrame may return \true 
when $Q$ is not homomorphic to $G$, 
but its embedding is slightly larger than that of $G$,
\ie the precision of \HFrame is low.
\looseness = -1

\etitle{Varying $d$}.
We varied the dimension $d$
of embeddings from 16 to 512.
We found the following.
(1) \HFrame gets slower as $d$ increases,
since the order-embedding space inspects
each dimension of embeddings;
and (2) the accuracy of \HFrame first increases
and then decreases, and its optimal performance achieves
when $d=64$.
This is because (a) when $d$ is small, 
multiple vertices can be mapped 
to the same embedding and 
the accuracy of \HFrame is low;
and (b) if $d$ is large, 
each embedding has many values,
and it is hard to ensure that the embedding 
of $Q$ is smaller than that of $G$,
even if $Q$ is homomorphic to $G$.
\looseness = -1

\etitle{Varying $\delta_{\rm avg}$}.
We generated synthetic graphs, 
and grouped them based on whether 
their average degrees $\delta_{\rm avg}$  are in the range $[d_1, d_2)$.
Each group contains 5000 graphs.
As shown in Table~\ref{tab:results-sensitive-0}, 
increasing graph density slightly reduces \HFrame's accuracy and increases its running time, as expected.
\looseness = -1

\stitle{Exp-4: Generalization}.
We next evaluated the generalization ability of \HFrame by training it on synthetic graphs 
and testing on real-life datasets.
As shown in Table~\ref{tab:generalization},
(1) \HFrame is robust to real-life patterns, 
achiving the accuracy above 0.860 on all graphs, 
which is lower than on synthetic data (0.968), but still outperforming all baselines.
(2) \HFrame trained on synthetic graphs performs comparably to models 
trained directly on real-life data, with only a 7.45\% drop in accuracy on average. 

\begin{table}[t]
		\caption{Generalization of \HFrame (Acc)}
	\label{tab:generalization}
	\vspace{-2ex}
	\centering
	\resizebox{1.0\columnwidth}{!}{
		\begin{tabular}{c|ccccccc}\hline
			\bf{Training Data}
			&Citation&Citeseerx&DBLP&IMDB&DBpedia&YAGO&Synthetic \\\hline
			Synthetic & 0.821 & 0.892 & 0.874 & 0872 & 0.853 & 0.824& 0.974 \\
			Real-life & 0.954 & 0.982 & 0.969 & 0.963 & 0.969 & 0.921 & -\\
			\hline
		\end{tabular}
	}
\end{table}

\stitle{Exp-5: Ablation Study}.
We conducted an ablation study
to evaluate the selection of framework design.
We compared \HFrame with the following variants:
(1) $\HFrame_{\kw{WS}}$, which omits \dual
for filtering candidates 
for each vertex in $Q$;
(2) $\HFrame_{\kw{MS}}$, which
uses multisets for message collection and aggregation; 
(3) $\HFrame_{\kw{WD}}$, that aggregates messages 
without distinguishing edge directions; 
(4) $\HFrame_{\kw{WN}}$, which 
skips normalization after computing embeddings; 
(5) $\HFrame_{\kw{WC}}$, 
which ignores cycle lengths in patterns when determine
the message-passing functions;
and (6) $\HFrame_{\kw{WG}}$, which
uses a loss function 
without  the $-\alpha$ term 
for the positive data and the term $\frac{1}{\kw{MSG}^{k,r}_1-\kw{MSG}^{k,r}_0}$. 

As showed in Tables~\ref{tab:accuracyresults}
and~\ref{tab:efficiency}, 
(1) \HFrame outperforms
all variants, as expected.
For example, on average \HFrame improves 
$\HFrame_{\kw{WS}}$, $\HFrame_{\kw{MS}}$, 
$\HFrame_{\kw{WD}}$, $\HFrame_{\kw{WN}}$, 
$\HFrame_{\kw{WC}}$ and $\HFrame_{\kw{WG}}$
by 2.74\%, 16.38\%, 6.38\%, 2.56\%, 3.92\% and 1.94\%, respectively.
(2) Filtering candidates 
with 
\dual and using sets to store received message
have the most impact on the performance of \HFrame,
since two distinct vertices in the pattern 
can be mapped to the  same vertex in the graph
via homomorphic mappings,
and collecting message with multisets results 
in multiple false negative results, which degrades
the performance of $\HFrame_{\kw{MS}}$.
(3) All variants have similar runtime,
since all optimizations
(\eg storing message using sets,
and normalizing the embeddings)
are efficient. 
\looseness = -1

\stitle{Exp-6: Applications}.
We show how to apply \HFrame in 
subgraph homomorphism matching and frequent pattern mining. 

\etitle{Subgraph homomorphism matching}.
We show how to enhance the performance of exact
matching algorithms
using \HFrame. 
More specifically, given a pattern $Q$ and a graph $G$, 
we (1) first invoke
\dual to filter candiates in ${\mathcal C}(u)$ for each pattern vertex $u$,
(2) next remove from $G$ all vertices 
that are not in ${\mathcal C}(u)$ for any $u$, 
(3) then use \HGIN to check 
whether there is a match for each vertex $v$ in ${\mathcal C}(u_1)$
of the first pattern vertex $u_1$ in the matching order ${\mathcal O}=(u_1, \ldots, u_n)$
of $Q$,
 and (4) finally, 
 run the exact algorithm to check 
 the remaining vertices in ${\mathcal C}(u_1)$.
 As shown in Table~\ref{tab:app-filter},
plugging \HFrame slightly 
reduces the accuracy of exact algorithm \kw{PJ},
but significantly accelerates its computation
on Citation and DBLP.

\begin{table}[t]
	\caption{Effectiveness of \HFrame in candidate filtering}
	\label{tab:app-filter}
	\vspace{-1.5ex}
	\centering
	\resizebox{0.8\columnwidth}{!}{
		\begin{tabular}{c|cc|cc} \hline
			\multirow{2}{*}{\bf{Method}}& \multicolumn{2}{c|}{\bf{F1 score}}
                & \multicolumn{2}{c}{\bf{Time(s)}}\\
			\cline{2-5}
					&Citation&DBLP&Citation&DBLP\\\hline
			\kw{PJ}              & 1.000 & 1.000 & 1.681 & 1.926 \\ 
			\kw{\HFrame+PJ}      & 0.832 & 0.835 & 0.762 & 0.780 \\ \hline
		\end{tabular}
	}
\end{table}

\etitle{Frequent pattern mining}. 
We demonstrate the application of \HFrame in frequent pattern mining.
Given a graph $G$,
frequent pattern mining is to identify patterns 
$Q$ with high support. 
The support of $Q$ is defined as the number of  vertices in $G$ that match the pivot (a designated vertex)
of $Q$.
We plug \HFrame in the frequent pattern mining algorithm SPMiner~\cite{ying2024representation},
	denoted by SPMiner+HF,
	to estimate the support of 
		the size-5 patterns that rank
	in the top 8 among all candiate patterns 
	by supports
	computed by the exact algorithm.
	On dataset Enzyme from~\cite{ying2024representation}, 
	SPMiner+HF
 computes the supports more accurately 
 than SPMiner,
 while maintaining  comparable  efficiency with SPMiner (159.35s vs.151.89s).
\looseness = -1

\begin{table}[tb!]
    \caption{Effectiveness of \HFrame in frequent pattern mining}
    \label{tab:app-mining}
    \vspace{-1.5ex}
    \centering
    \resizebox{1.01\columnwidth}{!}{
        \begin{tabular}{c|cccccccc} 
			\hline
            Method & 1 & 2 & 3 & 4 & 5 & 6 & 7 & 8 \\\hline 
            SPMiner    & 19,207 & 17,168 & 15,605 & 14,688 & 13,854 & 13,320 & 12,838 & 12,727 \\
            SPMiner+HF & 19,207 & 17,168 & 15,605 & 15,173 & 14,688 & 13,854 & 13,320 & 12,838 \\
            Exact      & 19,207 & 17,168 & 15,605 & 15,173 & 14,688 & 13,854 & 13,320 & 12,838\\ 
			\hline
        \end{tabular}
    }
\end{table}

\section{Conclusion}
In this paper, we proposed a framework \HFrame
for the subgraph homomorphism problem
by combining algorithmic solutions
and ML models.
We showed that \HFrame is more expressive
that the vanilla GNN for subgraph matching.
Moreover, we also provided the first generalization 
error
bound for the subgraph matching problem.
Using real-life and synthetic graphs, 
we experimentally validated the efficiency
and effectiveness of the proposed frameworks.
\looseness = -1

\begin{acks}
	This work was supported by the National Natural Science Foundation of China (No. U24B20143 and No. 62372030). 
\end{acks}

\balance
\bibliographystyle{ACM-Reference-Format}

\begin{thebibliography}{92}
	
\ifx \showCODEN    \undefined \def \showCODEN     #1{\unskip}     \fi
\ifx \showISBNx    \undefined \def \showISBNx     #1{\unskip}     \fi
\ifx \showISBNxiii \undefined \def \showISBNxiii  #1{\unskip}     \fi
\ifx \showISSN     \undefined \def \showISSN      #1{\unskip}     \fi
\ifx \showLCCN     \undefined \def \showLCCN      #1{\unskip}     \fi
\ifx \shownote     \undefined \def \shownote      #1{#1}          \fi
\ifx \showarticletitle \undefined \def \showarticletitle #1{#1}   \fi
\ifx \showURL      \undefined \def \showURL       {\relax}        \fi
\providecommand\bibfield[2]{#2}
\providecommand\bibinfo[2]{#2}
\providecommand\natexlab[1]{#1}
\providecommand\showeprint[2][]{arXiv:#2}

\bibitem[Cit(2021a)]%
{Citation}
\bibinfo{year}{2021}\natexlab{a}.
\newblock \bibinfo{title}{Citation}.
\newblock
\newblock
\shownote{{\sl https://aminer.cn/citation}}.


\bibitem[Cit(2021b)]%
{Citeseerx}
\bibinfo{year}{2021}\natexlab{b}.
\newblock \bibinfo{title}{Citeseerx}.
\newblock
\newblock
\shownote{{\sl citeseerx.ist.psu.edu}}.


\bibitem[DBL(2021)]%
{DBLP}
\bibinfo{year}{2021}\natexlab{}.
\newblock \bibinfo{title}{DBLP}.
\newblock
\newblock
\shownote{{\sl https://dblp.uni-trier.de/xml/}}.


\bibitem[DBp(2021)]%
{DBpedia}
\bibinfo{year}{2021}\natexlab{}.
\newblock \bibinfo{title}{DBpedia}.
\newblock
\newblock
\shownote{{\sl https://en.wikipedia.org/wiki/DBpedia}}.


\bibitem[IMD(2021)]%
{IMDB}
\bibinfo{year}{2021}\natexlab{}.
\newblock \bibinfo{title}{IMDB}.
\newblock
\newblock
\shownote{{\sl https://developer.imdb.com/non-commercial-datasets/}}.


\bibitem[YAG(2021)]%
{YAGO}
\bibinfo{year}{2021}\natexlab{}.
\newblock \bibinfo{title}{YAGO}.
\newblock
\newblock
\shownote{{\sl https://yago-knowledge.org/}}.


\bibitem[Neo(2023)]%
{Neo4j}
\bibinfo{year}{2023}\natexlab{}.
\newblock \bibinfo{title}{Neo4j 5.10.0}.
\newblock
\newblock
\shownote{{\sl https://neo4j.com/}}.


\bibitem[Spa(2025)]%
{Sparqlexample}
\bibinfo{year}{2025}\natexlab{}.
\newblock \bibinfo{title}{Wikidata: SPARQL query service/queries/examples}.
\newblock
\urldef\tempurl%
\\ \url{https://www.wikidata.org/wiki/Wikidata:SPARQL_query_service/queries/examples}
\showURL{%
	\tempurl}


\bibitem[Arai et~al\mbox{.}(2023)]%
{Iso-match-2}
\bibfield{author}{\bibinfo{person}{Junya Arai}, \bibinfo{person}{Yasuhiro
		Fujiwara}, {and} \bibinfo{person}{Makoto Onizuka}.}
\bibinfo{year}{2023}\natexlab{}.
\newblock \showarticletitle{GuP: Fast Subgraph Matching by Guard-based
	Pruning}.
\newblock \bibinfo{journal}{\emph{Proc. {ACM} Manag. Data}}
\bibinfo{volume}{1}, \bibinfo{number}{2} (\bibinfo{year}{2023}),
\bibinfo{pages}{167:1--167:26}.
\newblock


\bibitem[Barcel{\'{o}} et~al\mbox{.}(2022)]%
{barcelo2022weisfeiler}
\bibfield{author}{\bibinfo{person}{Pablo Barcel{\'{o}}},
	\bibinfo{person}{Mikhail Galkin}, \bibinfo{person}{Christopher Morris}, {and}
	\bibinfo{person}{Miguel A.~Romero Orth}.} \bibinfo{year}{2022}\natexlab{}.
\newblock \showarticletitle{Weisfeiler and leman go relational}. In
\bibinfo{booktitle}{\emph{{LoG}}}. \bibinfo{pages}{46--1}.
\newblock


\bibitem[Barcel{\'{o}} et~al\mbox{.}(2020)]%
{Expressive-FOC2}
\bibfield{author}{\bibinfo{person}{Pablo Barcel{\'{o}}},
	\bibinfo{person}{Egor~V. Kostylev}, \bibinfo{person}{Mika{\"{e}}l Monet},
	\bibinfo{person}{Jorge P{\'{e}}rez}, \bibinfo{person}{Juan~L. Reutter}, {and}
	\bibinfo{person}{Juan~Pablo Silva}.} \bibinfo{year}{2020}\natexlab{}.
\newblock \showarticletitle{The Expressive Power of Graph Neural Networks as a
	Query Language}.
\newblock \bibinfo{journal}{\emph{{SIGMOD} Rec.}} \bibinfo{volume}{49},
\bibinfo{number}{2} (\bibinfo{year}{2020}), \bibinfo{pages}{6--17}.
\newblock


\bibitem[Bhattarai et~al\mbox{.}(2019)]%
{CECI}
\bibfield{author}{\bibinfo{person}{Bibek Bhattarai}, \bibinfo{person}{Hang
		Liu}, {and} \bibinfo{person}{H.~Howie Huang}.}
\bibinfo{year}{2019}\natexlab{}.
\newblock \showarticletitle{{CECI:} Compact Embedding Cluster Index for
	Scalable Subgraph Matching}. In \bibinfo{booktitle}{\emph{{SIGMOD}}}.
\bibinfo{pages}{1447--1462}.
\newblock


\bibitem[Bi et~al\mbox{.}(2016)]%
{Iso-match-4}
\bibfield{author}{\bibinfo{person}{Fei Bi}, \bibinfo{person}{Lijun Chang},
	\bibinfo{person}{Xuemin Lin}, \bibinfo{person}{Lu Qin}, {and}
	\bibinfo{person}{Wenjie Zhang}.} \bibinfo{year}{2016}\natexlab{}.
\newblock \showarticletitle{Efficient Subgraph Matching by Postponing Cartesian
	Products}. In \bibinfo{booktitle}{\emph{{SIGMOD}}}.
\bibinfo{publisher}{{ACM}}, \bibinfo{pages}{1199--1214}.
\newblock


\bibitem[Bonnici et~al\mbox{.}(2013)]%
{Iso-match-5}
\bibfield{author}{\bibinfo{person}{Vincenzo Bonnici}, \bibinfo{person}{Rosalba
		Giugno}, \bibinfo{person}{Alfredo Pulvirenti}, \bibinfo{person}{Dennis~E.
		Shasha}, {and} \bibinfo{person}{Alfredo Ferro}.}
\bibinfo{year}{2013}\natexlab{}.
\newblock \showarticletitle{A subgraph isomorphism algorithm and its
	application to biochemical data}.
\newblock \bibinfo{journal}{\emph{{BMC} Bioinform.}} \bibinfo{volume}{14},
\bibinfo{number}{{S-7}} (\bibinfo{year}{2013}), \bibinfo{pages}{S13}.
\newblock


\bibitem[Bouritsas et~al\mbox{.}(2022)]%
{bouritsas2022improving}
\bibfield{author}{\bibinfo{person}{Giorgos Bouritsas},
	\bibinfo{person}{Fabrizio Frasca}, \bibinfo{person}{Stefanos Zafeiriou},
	{and} \bibinfo{person}{Michael~M Bronstein}.}
\bibinfo{year}{2022}\natexlab{}.
\newblock \showarticletitle{Improving graph neural network expressivity via
	subgraph isomorphism counting}.
\newblock \bibinfo{journal}{\emph{{IEEE Trans. Pattern Anal. Mach. Intell.}}}
\bibinfo{volume}{45}, \bibinfo{number}{1} (\bibinfo{year}{2022}),
\bibinfo{pages}{657--668}.
\newblock


\bibitem[Brynielsson et~al\mbox{.}(2010)]%
{brynielsson2010detecting}
\bibfield{author}{\bibinfo{person}{Joel Brynielsson}, \bibinfo{person}{Johanna
		H{\"o}gberg}, \bibinfo{person}{Lisa Kaati}, \bibinfo{person}{Christian
		M{\aa}rtenson}, {and} \bibinfo{person}{Pontus Svenson}.}
\bibinfo{year}{2010}\natexlab{}.
\newblock \showarticletitle{Detecting social positions using simulation}. In
\bibinfo{booktitle}{\emph{{ASONAM}}}. \bibinfo{pages}{48--55}.
\newblock


\bibitem[Cai et~al\mbox{.}(1992)]%
{kWL}
\bibfield{author}{\bibinfo{person}{Jin{-}yi Cai}, \bibinfo{person}{Martin
		F{\"{u}}rer}, {and} \bibinfo{person}{Neil Immerman}.}
\bibinfo{year}{1992}\natexlab{}.
\newblock \showarticletitle{An optimal lower bound on the number of variables
	for graph identification}.
\newblock \bibinfo{journal}{\emph{Comb.}} \bibinfo{volume}{12},
\bibinfo{number}{4} (\bibinfo{year}{1992}), \bibinfo{pages}{389--410}.
\newblock


\bibitem[Carletti et~al\mbox{.}(2019)]%
{carletti2019vf3}
\bibfield{author}{\bibinfo{person}{Vincenzo Carletti},
	\bibinfo{person}{Pasquale Foggia}, \bibinfo{person}{Antonio Greco},
	\bibinfo{person}{Mario Vento}, {and} \bibinfo{person}{Vincenzo Vigilante}.}
\bibinfo{year}{2019}\natexlab{}.
\newblock \showarticletitle{Vf3-light: A lightweight subgraph isomorphism
	algorithm and its experimental evaluation}.
\newblock \bibinfo{journal}{\emph{Pattern Recognition Letters}}
\bibinfo{volume}{125} (\bibinfo{year}{2019}), \bibinfo{pages}{591--596}.
\newblock


\bibitem[Carletti et~al\mbox{.}(2017)]%
{carletti2017introducing}
\bibfield{author}{\bibinfo{person}{Vincenzo Carletti},
	\bibinfo{person}{Pasquale Foggia}, \bibinfo{person}{Alessia Saggese}, {and}
	\bibinfo{person}{Mario Vento}.} \bibinfo{year}{2017}\natexlab{}.
\newblock \showarticletitle{Introducing VF3: A new algorithm for subgraph
	isomorphism}. In \bibinfo{booktitle}{\emph{{GbRPR}}}.
\bibinfo{pages}{128--139}.
\newblock


\bibitem[Chen et~al\mbox{.}(2019)]%
{chen2019hogmmnc}
\bibfield{author}{\bibinfo{person}{Jiazhou Chen}, \bibinfo{person}{Hong Peng},
	\bibinfo{person}{Guoqiang Han}, \bibinfo{person}{Hongmin Cai}, {and}
	\bibinfo{person}{Jiulun Cai}.} \bibinfo{year}{2019}\natexlab{}.
\newblock \showarticletitle{HOGMMNC: a higher order graph matching with
	multiple network constraints model for gene--drug regulatory modules
	identification}.
\newblock \bibinfo{journal}{\emph{Bioinformatics}} \bibinfo{volume}{35},
\bibinfo{number}{4} (\bibinfo{year}{2019}), \bibinfo{pages}{602--610}.
\newblock


\bibitem[Cheng et~al\mbox{.}(2010)]%
{cheng2010graph}
\bibfield{author}{\bibinfo{person}{Jiefeng Cheng}, \bibinfo{person}{Jeffrey~Xu
		Yu}, {and} \bibinfo{person}{S~Yu Philip}.} \bibinfo{year}{2010}\natexlab{}.
\newblock \showarticletitle{Graph pattern matching: A join/semijoin approach}.
\newblock \bibinfo{journal}{\emph{{TKDE}}} \bibinfo{volume}{23},
\bibinfo{number}{7} (\bibinfo{year}{2010}), \bibinfo{pages}{1006--1021}.
\newblock


\bibitem[Cheng et~al\mbox{.}(2013)]%
{cheng2013top}
\bibfield{author}{\bibinfo{person}{Jiefeng Cheng}, \bibinfo{person}{Xianggang
		Zeng}, {and} \bibinfo{person}{Jeffrey~Xu Yu}.}
\bibinfo{year}{2013}\natexlab{}.
\newblock \showarticletitle{Top-k graph pattern matching over large graphs}. In
\bibinfo{booktitle}{\emph{{ICDE}}}. \bibinfo{pages}{1033--1044}.
\newblock


\bibitem[Cook(2023)]%
{cook2023complexity}
\bibfield{author}{\bibinfo{person}{Stephen~A Cook}.}
\bibinfo{year}{2023}\natexlab{}.
\newblock \showarticletitle{The complexity of theorem-proving procedures}.
\newblock In \bibinfo{booktitle}{\emph{Logic, Automata, and Computational
		Complexity: The Works of Stephen A. Cook}}. \bibinfo{pages}{143--152}.
\newblock


\bibitem[Dragovic et~al\mbox{.}(2023)]%
{Cypher}
\bibfield{author}{\bibinfo{person}{Nikola Dragovic}, \bibinfo{person}{Cem
		Okulmus}, {and} \bibinfo{person}{Magdalena Ortiz}.}
\bibinfo{year}{2023}\natexlab{}.
\newblock \showarticletitle{Rewriting Ontology-Mediated Navigational Queries
	into Cypher}. In \bibinfo{booktitle}{\emph{{DL}}},
Vol.~\bibinfo{volume}{3515}.
\newblock


\bibitem[Du et~al\mbox{.}(2018)]%
{du2018personalized}
\bibfield{author}{\bibinfo{person}{Ruihuan Du}, \bibinfo{person}{Jiannan Yang},
	\bibinfo{person}{Yongzhi Cao}, {and} \bibinfo{person}{Hanpin Wang}.}
\bibinfo{year}{2018}\natexlab{}.
\newblock \showarticletitle{Personalized graph pattern matching via limited
	simulation}.
\newblock \bibinfo{journal}{\emph{Knowledge-Based Systems}}
\bibinfo{volume}{141} (\bibinfo{year}{2018}), \bibinfo{pages}{31--43}.
\newblock


\bibitem[Dutta et~al\mbox{.}(2017)]%
{dutta2017neighbor}
\bibfield{author}{\bibinfo{person}{Sourav Dutta}, \bibinfo{person}{Pratik
		Nayek}, {and} \bibinfo{person}{Arnab Bhattacharya}.}
\bibinfo{year}{2017}\natexlab{}.
\newblock \showarticletitle{Neighbor-aware search for approximate labeled graph
	matching using the chi-square statistics}. In
\bibinfo{booktitle}{\emph{{WWW}}}. \bibinfo{pages}{1281--1290}.
\newblock


\bibitem[Fan et~al\mbox{.}(2010a)]%
{fan2010grapha}
\bibfield{author}{\bibinfo{person}{Wenfei Fan}, \bibinfo{person}{Jianzhong Li},
	\bibinfo{person}{Shuai Ma}, \bibinfo{person}{Nan Tang},
	\bibinfo{person}{Yinghui Wu}, {and} \bibinfo{person}{Yunpeng Wu}.}
\bibinfo{year}{2010}\natexlab{a}.
\newblock \showarticletitle{Graph pattern matching: From intractable to
	polynomial time}.
\newblock \bibinfo{journal}{\emph{Proceedings of the VLDB Endowment}}
\bibinfo{volume}{3}, \bibinfo{number}{1-2} (\bibinfo{year}{2010}),
\bibinfo{pages}{264--275}.
\newblock


\bibitem[Fan et~al\mbox{.}(2010b)]%
{fan2010graph}
\bibfield{author}{\bibinfo{person}{Wenfei Fan}, \bibinfo{person}{Jianzhong Li},
	\bibinfo{person}{Shuai Ma}, \bibinfo{person}{Hongzhi Wang}, {and}
	\bibinfo{person}{Yinghui Wu}.} \bibinfo{year}{2010}\natexlab{b}.
\newblock \showarticletitle{Graph homomorphism revisited for graph matching}.
\newblock \bibinfo{journal}{\emph{{Proc. VLDB Endow.}}} \bibinfo{volume}{3},
\bibinfo{number}{1} (\bibinfo{year}{2010}), \bibinfo{pages}{1161--1172}.
\newblock


\bibitem[Fan et~al\mbox{.}(2024)]%
{Hercules}
\bibfield{author}{\bibinfo{person}{Wenfei Fan}, \bibinfo{person}{Kehan Pang},
	\bibinfo{person}{Ping Lu}, {and} \bibinfo{person}{Chao Tian}.}
\bibinfo{year}{2024}\natexlab{}.
\newblock \showarticletitle{Making It Tractable to Detect and Correct Errors in
	Graphs}.
\newblock \bibinfo{journal}{\emph{TODS}} \bibinfo{volume}{49},
\bibinfo{number}{4} (\bibinfo{year}{2024}), \bibinfo{pages}{16:1--16:75}.
\newblock


\bibitem[Fan et~al\mbox{.}(2013a)]%
{fei2013expfinder}
\bibfield{author}{\bibinfo{person}{Wenfei Fan}, \bibinfo{person}{Xin Wang},
	{and} \bibinfo{person}{Yinghui Wu}.} \bibinfo{year}{2013}\natexlab{a}.
\newblock \showarticletitle{ExpFinder: Finding experts by graph pattern
	matching}. In \bibinfo{booktitle}{\emph{{ICDE}}}.
\bibinfo{pages}{1316--1319}.
\newblock


\bibitem[Fan et~al\mbox{.}(2013b)]%
{fan2013incremental}
\bibfield{author}{\bibinfo{person}{Wenfei Fan}, \bibinfo{person}{Xin Wang},
	{and} \bibinfo{person}{Yinghui Wu}.} \bibinfo{year}{2013}\natexlab{b}.
\newblock \showarticletitle{Incremental graph pattern matching}.
\newblock \bibinfo{journal}{\emph{{TODS}}} \bibinfo{volume}{38},
\bibinfo{number}{3} (\bibinfo{year}{2013}), \bibinfo{pages}{1--47}.
\newblock


\bibitem[Fan et~al\mbox{.}(2014)]%
{fan2014distributed}
\bibfield{author}{\bibinfo{person}{Wenfei Fan}, \bibinfo{person}{Xin Wang},
	\bibinfo{person}{Yinghui Wu}, {and} \bibinfo{person}{Dong Deng}.}
\bibinfo{year}{2014}\natexlab{}.
\newblock \showarticletitle{Distributed graph simulation: Impossibility and
	possibility}.
\newblock \bibinfo{journal}{\emph{{Proc. VLDB Endow.}}} \bibinfo{volume}{7},
\bibinfo{number}{12} (\bibinfo{year}{2014}), \bibinfo{pages}{1083--1094}.
\newblock


\bibitem[Fard et~al\mbox{.}(2014)]%
{fard2014distributed}
\bibfield{author}{\bibinfo{person}{Arash Fard}, \bibinfo{person}{M~Usman
		Nisar}, \bibinfo{person}{John~A Miller}, {and} \bibinfo{person}{Lakshmish
		Ramaswamy}.} \bibinfo{year}{2014}\natexlab{}.
\newblock \showarticletitle{Distributed and scalable graph pattern matching:
	Models and algorithms}.
\newblock \bibinfo{journal}{\emph{{IJBD}}} \bibinfo{volume}{1},
\bibinfo{number}{1} (\bibinfo{year}{2014}), \bibinfo{pages}{1--14}.
\newblock


\bibitem[Feng et~al\mbox{.}(2022)]%
{feng2022powerful}
\bibfield{author}{\bibinfo{person}{Jiarui Feng}, \bibinfo{person}{Yixin Chen},
	\bibinfo{person}{Fuhai Li}, \bibinfo{person}{Anindya Sarkar}, {and}
	\bibinfo{person}{Muhan Zhang}.} \bibinfo{year}{2022}\natexlab{}.
\newblock \showarticletitle{How powerful are k-hop message passing graph neural
	networks}.
\newblock \bibinfo{journal}{\emph{{NeurIPS}}}  \bibinfo{volume}{35}
(\bibinfo{year}{2022}), \bibinfo{pages}{4776--4790}.
\newblock


\bibitem[Gao et~al\mbox{.}(2016)]%
{gao2016prs}
\bibfield{author}{\bibinfo{person}{Jianliang Gao}, \bibinfo{person}{Ping Liu},
	\bibinfo{person}{Xuedan Kang}, \bibinfo{person}{Lixia Zhang}, {and}
	\bibinfo{person}{Jianxin Wang}.} \bibinfo{year}{2016}\natexlab{}.
\newblock \showarticletitle{PRS: parallel relaxation simulation for massive
	graphs}.
\newblock \bibinfo{journal}{\emph{Comput. J.}} \bibinfo{volume}{59},
\bibinfo{number}{6} (\bibinfo{year}{2016}), \bibinfo{pages}{848--860}.
\newblock


\bibitem[Garg et~al\mbox{.}(2020)]%
{Gen-bound}
\bibfield{author}{\bibinfo{person}{Vikas~K. Garg}, \bibinfo{person}{Stefanie
		Jegelka}, {and} \bibinfo{person}{Tommi~S. Jaakkola}.}
\bibinfo{year}{2020}\natexlab{}.
\newblock \showarticletitle{Generalization and Representational Limits of Graph
	Neural Networks}. In \bibinfo{booktitle}{\emph{{ICML}}},
Vol.~\bibinfo{volume}{119}. \bibinfo{pages}{3419--3430}.
\newblock


\bibitem[Geerts and Reutter(2022)]%
{geerts2022expressiveness}
\bibfield{author}{\bibinfo{person}{Floris Geerts} {and}
	\bibinfo{person}{Juan~L. Reutter}.} \bibinfo{year}{2022}\natexlab{}.
\newblock \showarticletitle{Expressiveness and Approximation Properties of
	Graph Neural Networks}.
\newblock \bibinfo{journal}{\emph{{ICLR}}} (\bibinfo{year}{2022}).
\newblock


\bibitem[Geerts et~al\mbox{.}(2022)]%
{Express-GFMT}
\bibfield{author}{\bibinfo{person}{Floris Geerts}, \bibinfo{person}{Jasper
		Steegmans}, {and} \bibinfo{person}{Jan~Van den Bussche}.}
\bibinfo{year}{2022}\natexlab{}.
\newblock \showarticletitle{On the Expressive Power of Message-Passing Neural
	Networks as Global Feature Map Transformers}. In
\bibinfo{booktitle}{\emph{FoIKS}}, Vol.~\bibinfo{volume}{13388}.
\bibinfo{pages}{20--34}.
\newblock


\bibitem[Grohe(2023)]%
{Express-LICS}
\bibfield{author}{\bibinfo{person}{Martin Grohe}.}
\bibinfo{year}{2023}\natexlab{}.
\newblock \showarticletitle{The Descriptive Complexity of Graph Neural
	Networks}. In \bibinfo{booktitle}{\emph{{LICS}}}. \bibinfo{pages}{1--14}.
\newblock


\bibitem[Hamilton et~al\mbox{.}(2017)]%
{GraphSAGE}
\bibfield{author}{\bibinfo{person}{William~L. Hamilton},
	\bibinfo{person}{Zhitao Ying}, {and} \bibinfo{person}{Jure Leskovec}.}
\bibinfo{year}{2017}\natexlab{}.
\newblock \showarticletitle{Inductive Representation Learning on Large Graphs}.
In \bibinfo{booktitle}{\emph{NIPS}}. \bibinfo{pages}{1024--1034}.
\newblock


\bibitem[Han et~al\mbox{.}(2019)]%
{DpISO}
\bibfield{author}{\bibinfo{person}{Myoungji Han}, \bibinfo{person}{Hyunjoon
		Kim}, \bibinfo{person}{Geonmo Gu}, \bibinfo{person}{Kunsoo Park}, {and}
	\bibinfo{person}{Wook{-}Shin Han}.} \bibinfo{year}{2019}\natexlab{}.
\newblock \showarticletitle{Efficient Subgraph Matching: Harmonizing Dynamic
	Programming, Adaptive Matching Order, and Failing Set Together}. In
\bibinfo{booktitle}{\emph{{SIGMOD}}}. \bibinfo{pages}{1429--1446}.
\newblock


\bibitem[Han et~al\mbox{.}(2013)]%
{Iso-match-8}
\bibfield{author}{\bibinfo{person}{Wook{-}Shin Han}, \bibinfo{person}{Jinsoo
		Lee}, {and} \bibinfo{person}{Jeong{-}Hoon Lee}.}
\bibinfo{year}{2013}\natexlab{}.
\newblock \showarticletitle{Turbo\({}_{\mbox{iso}}\): towards ultrafast and
	robust subgraph isomorphism search in large graph databases}. In
\bibinfo{booktitle}{\emph{{SIGMOD}}}. \bibinfo{pages}{337--348}.
\newblock


\bibitem[Hang et~al\mbox{.}(2021)]%
{GNN-NC}
\bibfield{author}{\bibinfo{person}{Mengyue Hang}, \bibinfo{person}{Jennifer
		Neville}, {and} \bibinfo{person}{Bruno Ribeiro}.}
\bibinfo{year}{2021}\natexlab{}.
\newblock \showarticletitle{A Collective Learning Framework to Boost {GNN}
	Expressiveness for Node Classification}. In
\bibinfo{booktitle}{\emph{{ICML}}}, Vol.~\bibinfo{volume}{139}.
\bibinfo{pages}{4040--4050}.
\newblock


\bibitem[He and Singh(2008)]%
{Iso-match-7}
\bibfield{author}{\bibinfo{person}{Huahai He} {and} \bibinfo{person}{Ambuj~K.
		Singh}.} \bibinfo{year}{2008}\natexlab{}.
\newblock \showarticletitle{Graphs-at-a-time: query language and access methods
	for graph databases}. In \bibinfo{booktitle}{\emph{{SIGMOD}}}.
\bibinfo{publisher}{{ACM}}, \bibinfo{pages}{405--418}.
\newblock


\bibitem[Khan et~al\mbox{.}(2013)]%
{khan2013nema}
\bibfield{author}{\bibinfo{person}{Arijit Khan}, \bibinfo{person}{Yinghui Wu},
	\bibinfo{person}{Charu~C Aggarwal}, {and} \bibinfo{person}{Xifeng Yan}.}
\bibinfo{year}{2013}\natexlab{}.
\newblock \showarticletitle{Nema: Fast graph search with label similarity}.
\newblock \bibinfo{journal}{\emph{{Proc. VLDB Endow.}}} \bibinfo{volume}{6},
\bibinfo{number}{3} (\bibinfo{year}{2013}), \bibinfo{pages}{181--192}.
\newblock


\bibitem[Kim et~al\mbox{.}(2021)]%
{VEQ}
\bibfield{author}{\bibinfo{person}{Hyunjoon Kim}, \bibinfo{person}{Yunyoung
		Choi}, \bibinfo{person}{Kunsoo Park}, \bibinfo{person}{Xuemin Lin},
	\bibinfo{person}{Seok{-}Hee Hong}, {and} \bibinfo{person}{Wook{-}Shin Han}.}
\bibinfo{year}{2021}\natexlab{}.
\newblock \showarticletitle{Versatile Equivalences: Speeding up Subgraph Query
	Processing and Subgraph Matching}. In \bibinfo{booktitle}{\emph{{SIGMOD}}}.
\bibinfo{pages}{925--937}.
\newblock


\bibitem[Lan et~al\mbox{.}(2023)]%
{Sub-ML-4}
\bibfield{author}{\bibinfo{person}{Zixun Lan}, \bibinfo{person}{Limin Yu},
	\bibinfo{person}{Linglong Yuan}, \bibinfo{person}{Zili Wu},
	\bibinfo{person}{Qiang Niu}, {and} \bibinfo{person}{Fei Ma}.}
\bibinfo{year}{2023}\natexlab{}.
\newblock \showarticletitle{Sub-GMN: The Neural Subgraph Matching Network
	Model}. In \bibinfo{booktitle}{\emph{{CISP-BMEI}}}. \bibinfo{pages}{1--7}.
\newblock


\bibitem[Li et~al\mbox{.}(2022)]%
{li2022hismatch}
\bibfield{author}{\bibinfo{person}{Zixuan Li}, \bibinfo{person}{Zhongni Hou},
	\bibinfo{person}{Saiping Guan}, \bibinfo{person}{Xiaolong Jin},
	\bibinfo{person}{Weihua Peng}, \bibinfo{person}{Long Bai},
	\bibinfo{person}{Yajuan Lyu}, \bibinfo{person}{Wei Li},
	\bibinfo{person}{Jiafeng Guo}, {and} \bibinfo{person}{Xueqi Cheng}.}
\bibinfo{year}{2022}\natexlab{}.
\newblock \showarticletitle{Hismatch: Historical structure matching based
	temporal knowledge graph reasoning}.
\newblock \bibinfo{journal}{\emph{arXiv preprint arXiv:2210.09708}}
(\bibinfo{year}{2022}).
\newblock


\bibitem[Licheri et~al\mbox{.}(2021)]%
{licheri2021grapes}
\bibfield{author}{\bibinfo{person}{Nicola Licheri}, \bibinfo{person}{Vincenzo
		Bonnici}, \bibinfo{person}{Marco Beccuti}, {and} \bibinfo{person}{Rosalba
		Giugno}.} \bibinfo{year}{2021}\natexlab{}.
\newblock \showarticletitle{GRAPES-DD: exploiting decision diagrams for
	index-driven search in biological graph databases}.
\newblock \bibinfo{journal}{\emph{BMC bioinformatics}}  \bibinfo{volume}{22}
(\bibinfo{year}{2021}), \bibinfo{pages}{1--24}.
\newblock


\bibitem[Liu and Song(2022)]%
{DMPNN}
\bibfield{author}{\bibinfo{person}{Xin Liu} {and} \bibinfo{person}{Yangqiu
		Song}.} \bibinfo{year}{2022}\natexlab{}.
\newblock \showarticletitle{Graph Convolutional Networks with Dual Message
	Passing for Subgraph Isomorphism Counting and Matching}. In
\bibinfo{booktitle}{\emph{{AAAI}}}. \bibinfo{pages}{7594--7602}.
\newblock


\bibitem[Liu et~al\mbox{.}(2023)]%
{Sub-ML-1}
\bibfield{author}{\bibinfo{person}{Xuanzhou Liu}, \bibinfo{person}{Lin Zhang},
	\bibinfo{person}{Jiaqi Sun}, \bibinfo{person}{Yujiu Yang}, {and}
	\bibinfo{person}{Haiqin Yang}.} \bibinfo{year}{2023}\natexlab{}.
\newblock \showarticletitle{D2Match: Leveraging Deep Learning and Degeneracy
	for Subgraph Matching}. In \bibinfo{booktitle}{\emph{{ICML}}},
Vol.~\bibinfo{volume}{202}. \bibinfo{pages}{22454--22472}.
\newblock


\bibitem[Ma et~al\mbox{.}(2011)]%
{ma2011capturing}
\bibfield{author}{\bibinfo{person}{Shuai Ma}, \bibinfo{person}{Yang Cao},
	\bibinfo{person}{Wenfei Fan}, \bibinfo{person}{Jinpeng Huai}, {and}
	\bibinfo{person}{Tianyu Wo}.} \bibinfo{year}{2011}\natexlab{}.
\newblock \showarticletitle{Capturing topology in graph pattern matching}.
\newblock \bibinfo{journal}{\emph{Proc. {VLDB} Endow.}} \bibinfo{volume}{5},
\bibinfo{number}{4} (\bibinfo{year}{2011}), \bibinfo{pages}{310--321}.
\newblock


\bibitem[Ma et~al\mbox{.}(2014)]%
{StrongSimulation}
\bibfield{author}{\bibinfo{person}{Shuai Ma}, \bibinfo{person}{Yang Cao},
	\bibinfo{person}{Wenfei Fan}, \bibinfo{person}{Jinpeng Huai}, {and}
	\bibinfo{person}{Tianyu Wo}.} \bibinfo{year}{2014}\natexlab{}.
\newblock \showarticletitle{Strong simulation: Capturing topology in graph
	pattern matching}.
\newblock \bibinfo{journal}{\emph{TODS}} \bibinfo{volume}{39},
\bibinfo{number}{1} (\bibinfo{year}{2014}), \bibinfo{pages}{4:1--4:46}.
\newblock


\bibitem[Maron et~al\mbox{.}(2019)]%
{maron2018invariant}
\bibfield{author}{\bibinfo{person}{Haggai Maron}, \bibinfo{person}{Heli
		Ben-Hamu}, \bibinfo{person}{Nadav Shamir}, {and} \bibinfo{person}{Yaron
		Lipman}.} \bibinfo{year}{2019}\natexlab{}.
\newblock \showarticletitle{Invariant and equivariant graph networks}.
\newblock \bibinfo{journal}{\emph{{ICLR}}} (\bibinfo{year}{2019}).
\newblock


\bibitem[Morris et~al\mbox{.}(2023)]%
{VC-GNN}
\bibfield{author}{\bibinfo{person}{Christopher Morris}, \bibinfo{person}{Floris
		Geerts}, \bibinfo{person}{Jan T{\"{o}}nshoff}, {and} \bibinfo{person}{Martin
		Grohe}.} \bibinfo{year}{2023}\natexlab{}.
\newblock \showarticletitle{{WL} meet {VC}}. In
\bibinfo{booktitle}{\emph{{ICML}}}, Vol.~\bibinfo{volume}{202}.
\bibinfo{pages}{25275--25302}.
\newblock


\bibitem[Morris et~al\mbox{.}(2019)]%
{morris2019weisfeiler}
\bibfield{author}{\bibinfo{person}{Christopher Morris}, \bibinfo{person}{Martin
		Ritzert}, \bibinfo{person}{Matthias Fey}, \bibinfo{person}{William~L
		Hamilton}, \bibinfo{person}{Jan~Eric Lenssen}, \bibinfo{person}{Gaurav
		Rattan}, {and} \bibinfo{person}{Martin Grohe}.}
\bibinfo{year}{2019}\natexlab{}.
\newblock \showarticletitle{Weisfeiler and leman go neural: Higher-order graph
	neural networks}. In \bibinfo{booktitle}{\emph{{AAAI}}},
Vol.~\bibinfo{volume}{33}. \bibinfo{pages}{4602--4609}.
\newblock


\bibitem[Peng et~al\mbox{.}(2024)]%
{SPARQL}
\bibfield{author}{\bibinfo{person}{Peng Peng}, \bibinfo{person}{Shengyi Ji},
	\bibinfo{person}{M.~Tamer {\"{O}}zsu}, {and} \bibinfo{person}{Lei Zou}.}
\bibinfo{year}{2024}\natexlab{}.
\newblock \showarticletitle{Minimum motif-cut: a workload-aware {RDF} graph
	partitioning strategy}.
\newblock \bibinfo{journal}{\emph{{VLDB} J.}} \bibinfo{volume}{33},
\bibinfo{number}{5} (\bibinfo{year}{2024}), \bibinfo{pages}{1517--1542}.
\newblock


\bibitem[Roy et~al\mbox{.}(2022)]%
{Sub-ML-2}
\bibfield{author}{\bibinfo{person}{Indradyumna Roy}, \bibinfo{person}{Venkata
		Sai Baba~Reddy Velugoti}, \bibinfo{person}{Soumen Chakrabarti}, {and}
	\bibinfo{person}{Abir De}.} \bibinfo{year}{2022}\natexlab{}.
\newblock \showarticletitle{Interpretable Neural Subgraph Matching for Graph
	Retrieval}. In \bibinfo{booktitle}{\emph{{AAAI}}}.
\bibinfo{pages}{8115--8123}.
\newblock


\bibitem[Scarselli et~al\mbox{.}(2008)]%
{scarselli2008graph}
\bibfield{author}{\bibinfo{person}{Franco Scarselli}, \bibinfo{person}{Marco
		Gori}, \bibinfo{person}{Ah~Chung Tsoi}, \bibinfo{person}{Markus
		Hagenbuchner}, {and} \bibinfo{person}{Gabriele Monfardini}.}
\bibinfo{year}{2008}\natexlab{}.
\newblock \showarticletitle{The graph neural network model}.
\newblock \bibinfo{journal}{\emph{IEEE transactions on neural networks}}
\bibinfo{volume}{20}, \bibinfo{number}{1} (\bibinfo{year}{2008}),
\bibinfo{pages}{61--80}.
\newblock


\bibitem[Schlichtkrull et~al\mbox{.}(2018)]%
{r-GCN}
\bibfield{author}{\bibinfo{person}{Michael~Sejr Schlichtkrull},
	\bibinfo{person}{Thomas~N. Kipf}, \bibinfo{person}{Peter Bloem},
	\bibinfo{person}{Rianne van~den Berg}, \bibinfo{person}{Ivan Titov}, {and}
	\bibinfo{person}{Max Welling}.} \bibinfo{year}{2018}\natexlab{}.
\newblock \showarticletitle{Modeling Relational Data with Graph Convolutional
	Networks}. In \bibinfo{booktitle}{\emph{{ESWC}}},
Vol.~\bibinfo{volume}{10843}. \bibinfo{pages}{593--607}.
\newblock


\bibitem[Sharma et~al\mbox{.}(2024)]%
{GNN-recom}
\bibfield{author}{\bibinfo{person}{Kartik Sharma},
	\bibinfo{person}{Yeon{-}Chang Lee}, \bibinfo{person}{Sivagami Nambi},
	\bibinfo{person}{Aditya Salian}, \bibinfo{person}{Shlok Shah},
	\bibinfo{person}{Sang{-}Wook Kim}, {and} \bibinfo{person}{Srijan Kumar}.}
\bibinfo{year}{2024}\natexlab{}.
\newblock \showarticletitle{A Survey of Graph Neural Networks for Social
	Recommender Systems}.
\newblock \bibinfo{journal}{\emph{{ACM} Comput. Surv.}} \bibinfo{volume}{56},
\bibinfo{number}{10} (\bibinfo{year}{2024}), \bibinfo{pages}{265}.
\newblock


\bibitem[Shi et~al\mbox{.}(2017)]%
{shi2017multi}
\bibfield{author}{\bibinfo{person}{Qun Shi}, \bibinfo{person}{Guanfeng Liu},
	\bibinfo{person}{Kai Zheng}, \bibinfo{person}{An Liu}, \bibinfo{person}{Zhixu
		Li}, \bibinfo{person}{Lei Zhao}, {and} \bibinfo{person}{Xiaofang Zhou}.}
\bibinfo{year}{2017}\natexlab{}.
\newblock \showarticletitle{Multi-constrained top-K graph pattern matching in
	contextual social graphs}. In \bibinfo{booktitle}{\emph{{ICWS}}}.
\bibinfo{pages}{588--595}.
\newblock


\bibitem[Sun and Luo(2019)]%
{sun2019scaling}
\bibfield{author}{\bibinfo{person}{Shixuan Sun} {and} \bibinfo{person}{Qiong
		Luo}.} \bibinfo{year}{2019}\natexlab{}.
\newblock \showarticletitle{Scaling up subgraph query processing with efficient
	subgraph matching}. In \bibinfo{booktitle}{\emph{{ICDE}}}.
\bibinfo{pages}{220--231}.
\newblock


\bibitem[Sun and Luo(2020)]%
{sun2020subgraph}
\bibfield{author}{\bibinfo{person}{Shixuan Sun} {and} \bibinfo{person}{Qiong
		Luo}.} \bibinfo{year}{2020}\natexlab{}.
\newblock \showarticletitle{Subgraph matching with effective matching order and
	indexing}.
\newblock \bibinfo{journal}{\emph{{TKDE}}} \bibinfo{volume}{34},
\bibinfo{number}{1} (\bibinfo{year}{2020}), \bibinfo{pages}{491--505}.
\newblock


\bibitem[Sun et~al\mbox{.}(2020)]%
{RM}
\bibfield{author}{\bibinfo{person}{Shixuan Sun}, \bibinfo{person}{Xibo Sun},
	\bibinfo{person}{Yulin Che}, \bibinfo{person}{Qiong Luo}, {and}
	\bibinfo{person}{Bingsheng He}.} \bibinfo{year}{2020}\natexlab{}.
\newblock \showarticletitle{RapidMatch: {A} Holistic Approach to Subgraph Query
	Processing}.
\newblock \bibinfo{journal}{\emph{Proc. {VLDB} Endow.}} \bibinfo{volume}{14},
\bibinfo{number}{2} (\bibinfo{year}{2020}), \bibinfo{pages}{176--188}.
\newblock


\bibitem[Teru et~al\mbox{.}(2020)]%
{teru2020inductive}
\bibfield{author}{\bibinfo{person}{Komal Teru}, \bibinfo{person}{Etienne
		Denis}, {and} \bibinfo{person}{Will Hamilton}.}
\bibinfo{year}{2020}\natexlab{}.
\newblock \showarticletitle{Inductive relation prediction by subgraph
	reasoning}. In \bibinfo{booktitle}{\emph{International Conference on Machine
		Learning}}. PMLR, \bibinfo{pages}{9448--9457}.
\newblock


\bibitem[Tian et~al\mbox{.}(2007)]%
{tian2007saga}
\bibfield{author}{\bibinfo{person}{Yuanyuan Tian}, \bibinfo{person}{Richard~C
		Mceachin}, \bibinfo{person}{Carlos Santos}, \bibinfo{person}{David~J States},
	{and} \bibinfo{person}{Jignesh~M Patel}.} \bibinfo{year}{2007}\natexlab{}.
\newblock \showarticletitle{SAGA: a subgraph matching tool for biological
	graphs}.
\newblock \bibinfo{journal}{\emph{Bioinformatics}} \bibinfo{volume}{23},
\bibinfo{number}{2} (\bibinfo{year}{2007}), \bibinfo{pages}{232--239}.
\newblock


\bibitem[Ullmann(1976)]%
{Iso-match-6}
\bibfield{author}{\bibinfo{person}{Julian~R. Ullmann}.}
\bibinfo{year}{1976}\natexlab{}.
\newblock \showarticletitle{An Algorithm for Subgraph Isomorphism}.
\newblock \bibinfo{journal}{\emph{J. {ACM}}} \bibinfo{volume}{23},
\bibinfo{number}{1} (\bibinfo{year}{1976}), \bibinfo{pages}{31--42}.
\newblock


\bibitem[Vendrov et~al\mbox{.}(2016)]%
{Order-embedding-1}
\bibfield{author}{\bibinfo{person}{Ivan Vendrov}, \bibinfo{person}{Ryan Kiros},
	\bibinfo{person}{Sanja Fidler}, {and} \bibinfo{person}{Raquel Urtasun}.}
\bibinfo{year}{2016}\natexlab{}.
\newblock \showarticletitle{Order-Embeddings of Images and Language}. In
\bibinfo{booktitle}{\emph{{ICLR}}}.
\newblock


\bibitem[Wang et~al\mbox{.}(2022b)]%
{wang2022equivariant}
\bibfield{author}{\bibinfo{person}{Haorui Wang}, \bibinfo{person}{Haoteng Yin},
	\bibinfo{person}{Muhan Zhang}, {and} \bibinfo{person}{Pan Li}.}
\bibinfo{year}{2022}\natexlab{b}.
\newblock \showarticletitle{Equivariant and stable positional encoding for more
	powerful graph neural networks}.
\newblock \bibinfo{journal}{\emph{{ICLR}}} (\bibinfo{year}{2022}).
\newblock


\bibitem[Wang et~al\mbox{.}(2021)]%
{GNN-AD}
\bibfield{author}{\bibinfo{person}{Yanling Wang}, \bibinfo{person}{Jing Zhang},
	\bibinfo{person}{Shasha Guo}, \bibinfo{person}{Hongzhi Yin},
	\bibinfo{person}{Cuiping Li}, {and} \bibinfo{person}{Hong Chen}.}
\bibinfo{year}{2021}\natexlab{}.
\newblock \showarticletitle{Decoupling Representation Learning and
	Classification for GNN-based Anomaly Detection}. In
\bibinfo{booktitle}{\emph{{SIGIR}}}. \bibinfo{pages}{1239--1248}.
\newblock


\bibitem[Wang et~al\mbox{.}(2022a)]%
{wang2022twin}
\bibfield{author}{\bibinfo{person}{Zhaohui Wang}, \bibinfo{person}{Qi Cao},
	\bibinfo{person}{Huawei Shen}, \bibinfo{person}{Bingbing Xu}, {and}
	\bibinfo{person}{Xueqi Cheng}.} \bibinfo{year}{2022}\natexlab{a}.
\newblock \showarticletitle{Twin Weisfeiler-Lehman: High Expressive GNNs for
	Graph Classification}.
\newblock \bibinfo{journal}{\emph{arXiv preprint arXiv:2203.11683}}
(\bibinfo{year}{2022}).
\newblock


\bibitem[Wang et~al\mbox{.}(2024)]%
{Generalization-gap}
\bibfield{author}{\bibinfo{person}{Zhiyang Wang}, \bibinfo{person}{Juan
		Cervi{\~{n}}o}, {and} \bibinfo{person}{Alejandro Ribeiro}.}
\bibinfo{year}{2024}\natexlab{}.
\newblock \showarticletitle{Generalization of Geometric Graph Neural Networks}.
\newblock \bibinfo{journal}{\emph{CoRR}}  \bibinfo{volume}{abs/2409.05191}
(\bibinfo{year}{2024}).
\newblock


\bibitem[Weisfeiler and Leman(1968)]%
{1WL}
\bibfield{author}{\bibinfo{person}{Boris Weisfeiler} {and}
	\bibinfo{person}{Andrei Leman}.} \bibinfo{year}{1968}\natexlab{}.
\newblock \showarticletitle{The reduction of a graph to canonical form and the
	algebra which appears therein}.
\newblock \bibinfo{journal}{\emph{nti, Series}} \bibinfo{volume}{2},
\bibinfo{number}{9} (\bibinfo{year}{1968}), \bibinfo{pages}{12--16}.
\newblock


\bibitem[Wu et~al\mbox{.}(2023)]%
{wu2023novel}
\bibfield{author}{\bibinfo{person}{Xiaoying Wu}, \bibinfo{person}{Dimitri
		Theodoratos}, \bibinfo{person}{Dimitrios Skoutas}, {and}
	\bibinfo{person}{Michael Lan}.} \bibinfo{year}{2023}\natexlab{}.
\newblock \showarticletitle{A novel framework for the efficient evaluation of
	hybrid tree-pattern queries on large data graphs}.
\newblock \bibinfo{journal}{\emph{Information Systems}}  \bibinfo{volume}{117}
(\bibinfo{year}{2023}), \bibinfo{pages}{102249}.
\newblock


\bibitem[Wu et~al\mbox{.}(2013)]%
{wu2013ontology}
\bibfield{author}{\bibinfo{person}{Yinghui Wu}, \bibinfo{person}{Shengqi Yang},
	{and} \bibinfo{person}{Xifeng Yan}.} \bibinfo{year}{2013}\natexlab{}.
\newblock \showarticletitle{Ontology-based subgraph querying}. In
\bibinfo{booktitle}{\emph{{ICDE}}}. \bibinfo{pages}{697--708}.
\newblock


\bibitem[Xu et~al\mbox{.}(2019a)]%
{xu2018powerful}
\bibfield{author}{\bibinfo{person}{Keyulu Xu}, \bibinfo{person}{Weihua Hu},
	\bibinfo{person}{Jure Leskovec}, {and} \bibinfo{person}{Stefanie Jegelka}.}
\bibinfo{year}{2019}\natexlab{a}.
\newblock \showarticletitle{How powerful are graph neural networks?}
\newblock \bibinfo{journal}{\emph{{ICLR}}} (\bibinfo{year}{2019}).
\newblock


\bibitem[Xu et~al\mbox{.}(2019b)]%
{xu2019cross}
\bibfield{author}{\bibinfo{person}{Kun Xu}, \bibinfo{person}{Liwei Wang},
	\bibinfo{person}{Mo Yu}, \bibinfo{person}{Yansong Feng}, \bibinfo{person}{Yan
		Song}, \bibinfo{person}{Zhiguo Wang}, {and} \bibinfo{person}{Dong Yu}.}
\bibinfo{year}{2019}\natexlab{b}.
\newblock \showarticletitle{Cross-lingual knowledge graph alignment via graph
	matching neural network}.
\newblock \bibinfo{journal}{\emph{arXiv preprint arXiv:1905.11605}}
(\bibinfo{year}{2019}).
\newblock


\bibitem[Yan et~al\mbox{.}(2004)]%
{yan2004graph}
\bibfield{author}{\bibinfo{person}{Xifeng Yan}, \bibinfo{person}{Philip~S Yu},
	{and} \bibinfo{person}{Jiawei Han}.} \bibinfo{year}{2004}\natexlab{}.
\newblock \showarticletitle{Graph indexing: a frequent structure-based
	approach}. In \bibinfo{booktitle}{\emph{{SIGMOD}}}.
\bibinfo{pages}{335--346}.
\newblock


\bibitem[Yang et~al\mbox{.}(2023)]%
{Iso-match-3}
\bibfield{author}{\bibinfo{person}{Rongjian Yang}, \bibinfo{person}{Zhijie
		Zhang}, \bibinfo{person}{Weiguo Zheng}, {and} \bibinfo{person}{Jeffrey~Xu
		Yu}.} \bibinfo{year}{2023}\natexlab{}.
\newblock \showarticletitle{Fast Continuous Subgraph Matching over Streaming
	Graphs via Backtracking Reduction}.
\newblock \bibinfo{journal}{\emph{Proc. {ACM} Manag. Data}}
\bibinfo{volume}{1}, \bibinfo{number}{1} (\bibinfo{year}{2023}),
\bibinfo{pages}{15:1--15:26}.
\newblock


\bibitem[Ye et~al\mbox{.}(2024)]%
{Sub-ML-3}
\bibfield{author}{\bibinfo{person}{Yutong Ye}, \bibinfo{person}{Xiang Lian},
	{and} \bibinfo{person}{Mingsong Chen}.} \bibinfo{year}{2024}\natexlab{}.
\newblock \showarticletitle{Efficient Exact Subgraph Matching via GNN-based
	Path Dominance Embedding}.
\newblock \bibinfo{journal}{\emph{Proc. {VLDB} Endow.}} \bibinfo{volume}{17},
\bibinfo{number}{7} (\bibinfo{year}{2024}), \bibinfo{pages}{1628--1641}.
\newblock


\bibitem[Ying et~al\mbox{.}(2024)]%
{ying2024representation}
\bibfield{author}{\bibinfo{person}{Rex Ying}, \bibinfo{person}{Tianyu Fu},
	\bibinfo{person}{Andrew Wang}, \bibinfo{person}{Jiaxuan You},
	\bibinfo{person}{Yu Wang}, {and} \bibinfo{person}{Jure Leskovec}.}
\bibinfo{year}{2024}\natexlab{}.
\newblock \showarticletitle{Representation Learning for Frequent Subgraph
	Mining}.
\newblock \bibinfo{journal}{\emph{arXiv preprint arXiv:2402.14367}}
(\bibinfo{year}{2024}).
\newblock


\bibitem[Ying et~al\mbox{.}(2020)]%
{NeuroMatch}
\bibfield{author}{\bibinfo{person}{Rex Ying}, \bibinfo{person}{Zhaoyu Lou},
	\bibinfo{person}{Jiaxuan You}, \bibinfo{person}{Chengtao Wen},
	\bibinfo{person}{Arquimedes Canedo}, {and} \bibinfo{person}{Jure Leskovec}.}
\bibinfo{year}{2020}\natexlab{}.
\newblock \showarticletitle{Neural Subgraph Matching}.
\newblock \bibinfo{journal}{\emph{CoRR}}  \bibinfo{volume}{abs/2007.03092}
(\bibinfo{year}{2020}).
\newblock
\urldef\tempurl%
\url{https://arxiv.org/abs/2007.03092}
\showURL{%
	\tempurl}


\bibitem[You et~al\mbox{.}(2021)]%
{you2021identity}
\bibfield{author}{\bibinfo{person}{Jiaxuan You}, \bibinfo{person}{Jonathan~M
		Gomes-Selman}, \bibinfo{person}{Rex Ying}, {and} \bibinfo{person}{Jure
		Leskovec}.} \bibinfo{year}{2021}\natexlab{}.
\newblock \showarticletitle{Identity-aware graph neural networks}. In
\bibinfo{booktitle}{\emph{{AAAI}}}, Vol.~\bibinfo{volume}{35}.
\bibinfo{pages}{10737--10745}.
\newblock


\bibitem[Yuan et~al\mbox{.}(2011)]%
{yuan2011efficient}
\bibfield{author}{\bibinfo{person}{Ye Yuan}, \bibinfo{person}{Guoren Wang},
	\bibinfo{person}{Haixun Wang}, {and} \bibinfo{person}{Lei Chen}.}
\bibinfo{year}{2011}\natexlab{}.
\newblock \showarticletitle{Efficient subgraph search over large uncertain
	graphs}.
\newblock \bibinfo{journal}{\emph{Proc. VLDB Endow.}} \bibinfo{volume}{4},
\bibinfo{number}{11} (\bibinfo{year}{2011}), \bibinfo{pages}{876--886}.
\newblock


\bibitem[Zhang et~al\mbox{.}(2023a)]%
{zhang2023expressive}
\bibfield{author}{\bibinfo{person}{Bingxu Zhang}, \bibinfo{person}{Changjun
		Fan}, \bibinfo{person}{Shixuan Liu}, \bibinfo{person}{Kuihua Huang},
	\bibinfo{person}{Xiang Zhao}, \bibinfo{person}{Jincai Huang}, {and}
	\bibinfo{person}{Zhong Liu}.} \bibinfo{year}{2023}\natexlab{a}.
\newblock \showarticletitle{The expressive power of graph neural networks: A
	survey}.
\newblock \bibinfo{journal}{\emph{arXiv preprint arXiv:2308.08235}}
(\bibinfo{year}{2023}).
\newblock


\bibitem[Zhang et~al\mbox{.}(2023b)]%
{GNN-WL-C}
\bibfield{author}{\bibinfo{person}{Bohang Zhang}, \bibinfo{person}{Shengjie
		Luo}, \bibinfo{person}{Liwei Wang}, {and} \bibinfo{person}{Di He}.}
\bibinfo{year}{2023}\natexlab{b}.
\newblock \showarticletitle{Rethinking the Expressive Power of GNNs via Graph
	Biconnectivity}. In \bibinfo{booktitle}{\emph{{ICLR}}}.
\newblock


\bibitem[Zhang et~al\mbox{.}(2024a)]%
{Simulation-2}
\bibfield{author}{\bibinfo{person}{Tianming Zhang}, \bibinfo{person}{Xinwei
		Cai}, \bibinfo{person}{Lu Chen}, \bibinfo{person}{Zhengyi Yang},
	\bibinfo{person}{Yunjun Gao}, \bibinfo{person}{Bin Cao}, {and}
	\bibinfo{person}{Jing Fan}.} \bibinfo{year}{2024}\natexlab{a}.
\newblock \showarticletitle{Towards efficient simulation-based constrained
	temporal graph pattern matching}.
\newblock \bibinfo{journal}{\emph{World Wide Web {(WWW)}}}
\bibinfo{volume}{27}, \bibinfo{number}{3} (\bibinfo{year}{2024}),
\bibinfo{pages}{22}.
\newblock


\bibitem[Zhang et~al\mbox{.}(2021)]%
{zhang2021eigen}
\bibfield{author}{\bibinfo{person}{Ziwei Zhang}, \bibinfo{person}{Peng Cui},
	\bibinfo{person}{Jian Pei}, \bibinfo{person}{Xin Wang}, {and}
	\bibinfo{person}{Wenwu Zhu}.} \bibinfo{year}{2021}\natexlab{}.
\newblock \showarticletitle{Eigen-gnn: A graph structure preserving plug-in for
	gnns}.
\newblock \bibinfo{journal}{\emph{{TKDE}}} \bibinfo{volume}{35},
\bibinfo{number}{3} (\bibinfo{year}{2021}), \bibinfo{pages}{2544--2555}.
\newblock


\bibitem[Zhang et~al\mbox{.}(2024b)]%
{Survey-matching}
\bibfield{author}{\bibinfo{person}{Zhijie Zhang}, \bibinfo{person}{Yujie Lu},
	\bibinfo{person}{Weiguo Zheng}, {and} \bibinfo{person}{Xuemin Lin}.}
\bibinfo{year}{2024}\natexlab{b}.
\newblock \showarticletitle{A Comprehensive Survey and Experimental Study of
	Subgraph Matching: Trends, Unbiasedness, and Interaction}.
\newblock \bibinfo{journal}{\emph{Proc. {ACM} Manag. Data}}
\bibinfo{volume}{2}, \bibinfo{number}{1} (\bibinfo{year}{2024}),
\bibinfo{pages}{60:1--60:29}.
\newblock


\bibitem[Zhao et~al\mbox{.}(2007)]%
{zhao2007graph}
\bibfield{author}{\bibinfo{person}{Peixiang Zhao}, \bibinfo{person}{Jeffrey~Xu
		Yu}, {and} \bibinfo{person}{Philip~S Yu}.} \bibinfo{year}{2007}\natexlab{}.
\newblock \showarticletitle{Graph indexing: tree + delta $>$= graph}. In
\bibinfo{booktitle}{\emph{VLDB}}. \bibinfo{pages}{938--949}.
\newblock


\bibitem[Zhao et~al\mbox{.}(2013)]%
{zhao2013partition}
\bibfield{author}{\bibinfo{person}{Xiang Zhao}, \bibinfo{person}{Chuan Xiao},
	\bibinfo{person}{Xuemin Lin}, \bibinfo{person}{Qing Liu}, {and}
	\bibinfo{person}{Wenjie Zhang}.} \bibinfo{year}{2013}\natexlab{}.
\newblock \showarticletitle{A partition-based approach to structure similarity
	search}.
\newblock \bibinfo{journal}{\emph{{Proc. VLDB Endow.}}} \bibinfo{volume}{7},
\bibinfo{number}{3} (\bibinfo{year}{2013}), \bibinfo{pages}{169--180}.
\newblock
        
\end{thebibliography}

\balance
\appendix

\section{Appendix}

\begin{figure*}[h]
	\begin{center}
		\begin{minipage}[t]{0.6\textwidth}
			\subfigure{{\includegraphics[width=\columnwidth]{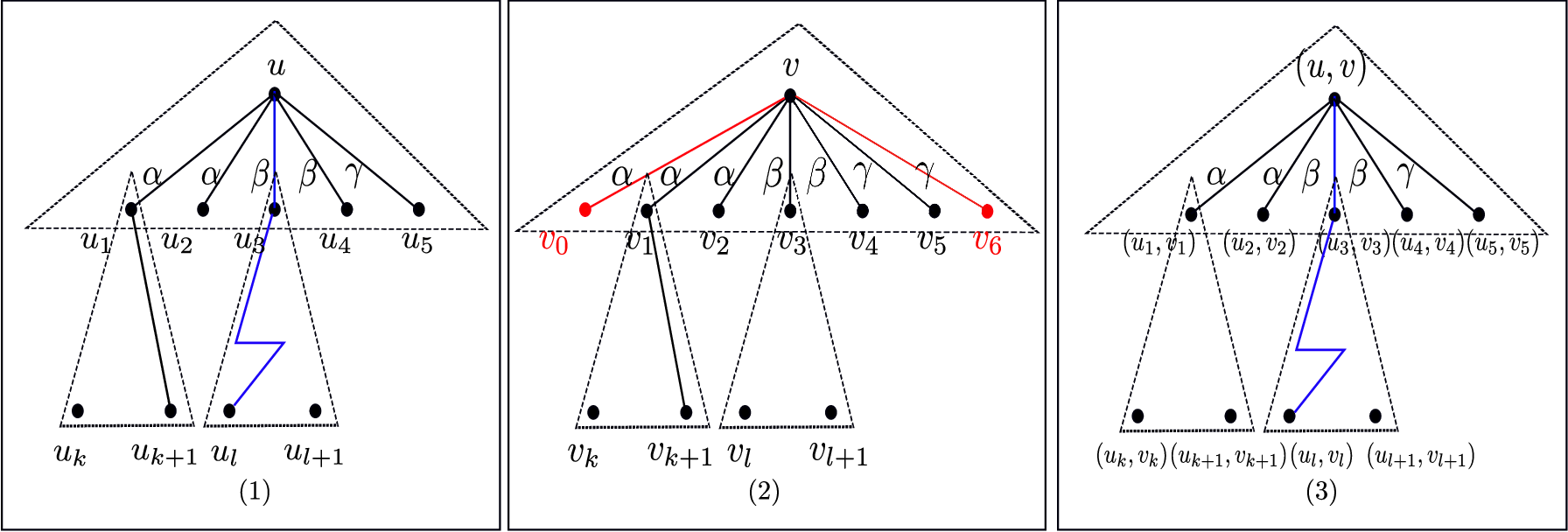}}}
			\hfill
		\end{minipage}
	\end{center}
	\vspace{-4ex}
	\caption{Computation of embeddings and \dual} 
	\label{fig:anti-expressive-0}
\end{figure*}

\subsection{Proof of Theorem~\ref{thm-upper-bound}}
Assume by contradiction 
that there  exist a pattern $Q$, a graph $G$,
a vertex $u$ in $Q$ and a vertex $v$ in $G$ 
such that $\HGIN^\kw{v}$ returns \false, 
but \dual returns \true, \ie $(u, v)$ exists
in the relation $S_{(Q, G)}$ computed by \dual.
We deduce a contradiction by analyzing 
the computation of $\HGIN^\kw{v}$.
Observe that
(a) computing
embeddings of $u$ and $v$ can be represented by
trees of height $m$, where $m$ is 
the number of layers in $\HGIN^\kw{v}$
(\cite{Gen-bound}; see Figures~\ref{fig:anti-expressive-0}(a) and~\ref{fig:anti-expressive-0}(b)
for an example);
and (b) vertex $v$ may have more 
adjacent edges in $G$ than $u$, 
\eg paths colored in red in Figure~\ref{fig:anti-expressive-0}(b).
When $\HGIN^\kw{v}$ returns \false, 
the embedding of $v$ is not larger than that of $u$
(see Section~\ref{sec-model}).
Because all functions $\kw{MSG}^{k,r}$ and $\kw{AGG}^k$
$(k\in \mathcal{L})$ in $\HGIN^v$
are injective and monotonic,
there exists a path $p_0$ in the tree representing the computation of 
embedding of $u$ (\eg the path colored in blue
in Figure~\ref{fig:anti-expressive-0}(a)), such that 
no path from $v$ carries the same labels as $p_0$.
However, if this holds, \dual can also identify 
such path and returns \false.
Indeed, the computation of \dual can also be represented
as a tree (see Figures~\ref{fig:anti-expressive-0}(c)
for an example).
The match relation $S_{(Q, G)}$
can be constructed from common parts 
of trees roots at $u$ and $v$ (see Section~\ref{sec-pre}).
When such a path $p_0$ exists in $G$, \dual can also identify 
that $p_0$ is missed in the tree of $v$, 
and $(u, v)$ is not contained in $S_{(Q, G)}$, 
\ie~\dual returns \false, a contradiction.
\looseness = -1

\subsection{Proof of Theorem~\ref{thm-hgin-dual}}
(1) We first show that given a pattern $Q$, a graph $G$,
a vertex $u$ in $Q$ and a vertex $v$ in $G$,
if $(u, v)\not\in S_{(Q, G)}$,
then there exists a $\HGIN$ that returns \false
given the input.
Indeed, when $(u, v)\not\in S_{(Q, G)}$,
there exists a path $p$ from $u$ in $Q$ such that 
$p$ is not a path from $v$ in $G$.
Then we can construct a $\HGIN$ 
such that it has $|p|$ layers,
and all its functions $\kw{MSG}^{k,r}$
and $\kw{AGG}^k$
$(k\leq |p|)$
are injective and monotonic.
Then the embedding 
of $u$ computed by $\HGIN$ 
is not smaller than that of $v$, 
\ie $\HGIN$  returns \false, 
since the embedding of 
$p$ is used to compute the embedding of $u$ but not in that of $v$, 
and functions $\kw{MSG}^{k,r}$
and $\kw{AGG}^k$
$(k\leq |p|)$ in $\HGIN$
are injective and monotonic.

\begin{figure}[th!]
	\begin{center}
		\begin{tabular}{c}
			\includegraphics[width=0.6\columnwidth]{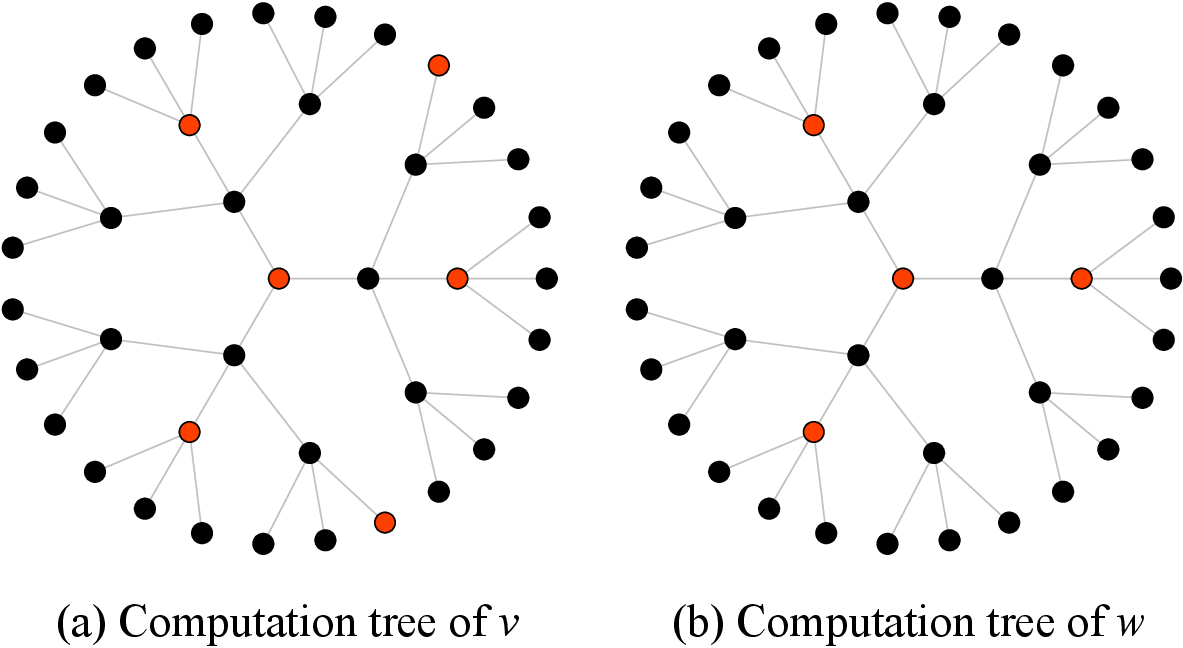}
		\end{tabular}
	\end{center}
	\vspace{-3ex}
	\caption{Computation trees of embeddings} 
	\label{fig:express-4}
\end{figure}

\stab
(2) 
We next show that 
$\HGIN$ is more expressive than
\dual.
Consider graphs $G_2$ and $G_3$ in Figure~\ref{fig:difference}, 
which are extended from~\cite{you2021identity}.
We can verify that 
there does not exist a homomorphic mapping 
from $G_3$ to $G_2$, since $G_3$ contains
triangles (\ie the subgraph induced by 
$w_1, w_5$ and $w_6$), while $G_2$ does not.
We next show that
given $G_3$, $G_2$, $v_1$
and $w_1$,
(1) \HGIN returns \false 
but (2) $(v_1, w_1){\in} S_{(G_3,G_2)}$.
If these hold, then 
\HGIN is more expressive than \dual,
{\em i.e.}, \HGIN can distinct more non-homomorphic
graphs.
\looseness=-1

\stab
(I) We first prove that \HGIN returns \false.
This is because (a) $v_1$ is contained in a cycle ${\mathcal C}$ of length 3, 
while $w_1$ does not;
(b) \HGIN ensures that $\kw{MSG}^{k,r}_1$ is sufficiently 
larger than $\kw{MSG}^{k,r}_0$; and
(c) the cycle ${\mathcal C}$ ensures that 
computing the embedding of $v_1$ 
invokes more $\kw{MSG}^{k,r}_1$, 
which leads to the embedding
of $v_1$ being larger than that of $w_1$.
Consider two trees in Figure~\ref{fig:express-4}, 
which represent the computation of embeddings of 
$v_1$ and $w_1$. The vertices with message passing function $\kw{MSG}^{k,r}_1$ are marked red.
We can see that the computation of embedding of $v_1$ invokes more times of  function 
$\kw{MSG}^{k,r}_1$ than the computation of embedding of $w_1$.
\looseness = -1

\stab	
(II) We next show that 
$(v_1, w_1)\in S_{(G_3,G_2)}$.
That is, $(v_1, w_1)$ 
exists in the dual simulation of $G_3$ and $G_2$.
Observe that both  $G_3$ and $G_2$ are regular, 
and all vertices have the same number of
neighbors.
Therefore, all vertices pairs 
exist in $V_{G_3}\times V_{G_2}$,
\ie $S_{(G_3, G_2)}=V_{G_3}{\times} V_{G_2}$.

\begin{figure}[th!]
	\centerline{\includegraphics[width=0.8\columnwidth]{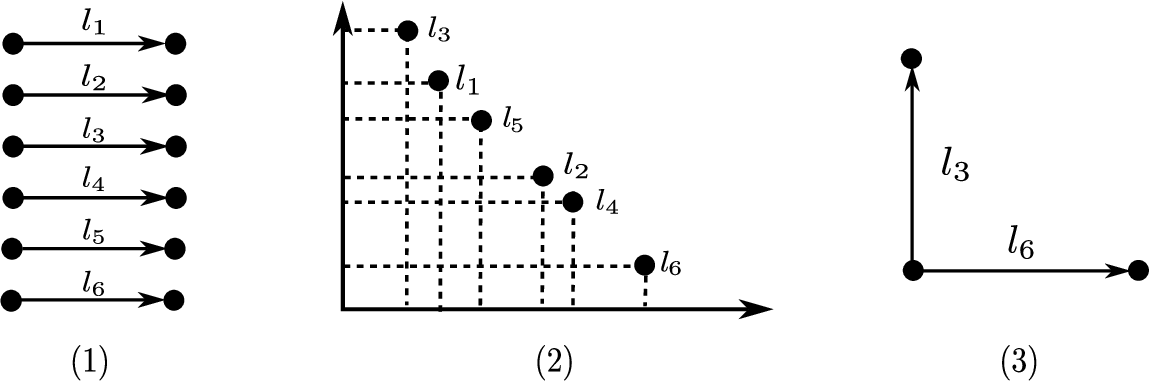}}
	\centering
	\caption{Example in Theorem~\ref{thm-more-expressive-1}} 
	\label{fig:label-embedding}
\end{figure}

\subsection*{A3. Proof of Theorem~\ref{thm-more-expressive-1}}

Assume that the embeddings are finite dimension, 
and let the embedding be $d$-dimension.
We select $2d$ distinct edge labels
$l_1, \ldots, l_{2d}$, 
which embeddings are $e_1, \ldots, e_{2d}$, respectively.
Since the embeddings are $d$-dimension, 
we can pick $d$ embeddings $e'_1, \ldots, e'_d$
such that the $i$-th index of $e'_i$ 
is maximum among all $2d$ embeddings $e_1, \ldots, e_{2d}$.
Let $l'_1, \ldots, l'_d$ be the labels with embeddings
 $e'_1, \ldots, e'_d$, respectively.
Next we construct a star pattern $Q_s$ 
with $d$ edges adjacent to the same vertex, 
where the edge  labels 
are the picked $d$ labels $l'_1, \ldots, l'_d$.
We can deduce a contradiction by comparing
the embeddings of $Q_s$ and 
that of all these $2d$ labels.
(1) Because $Q_s$ consists of $d$ edge labels, 
the embedding of $Q_s$ must be larger than 
the embeddings of these $d$ labels, based on the 
order-embedding space (see the prediction function
of \HGIN);
(2) each of these $d$ vectors 
has the maximum value for at least one index,
then the embedding of $Q_s$ must be larger than 
the embeddings of all $2d$ labels,
which is a contradiction, 
since $Q_s$ has only $d$ labels,
and $l_1, \ldots, l_{2d}$ are distinct.

Consider the six edges in Figure~\ref{fig:label-embedding}(1),
which carry distinct labels
$l_1, \ldots, l_6$. Their embeddings in two-dimensional space
as shown in Figure~\ref{fig:label-embedding}(b).
Since these labels are distinct, 
none of these labels is homomorphic 
to any of the other labels.
Therefore, given any embeddings
$e_1 = (a_1, a_2)$ and 
$e_2 = (b_1, b_2)$ of two labels 
$l_1$ and $l_2$, respectively,
if $a_1\leq b_1$, then $a_2> b_2$;
and if $a_2\leq b_2$, then $a_1> b_1$ 
(see Figure~\ref{fig:label-embedding}(2)).
Then we pick two labels 
$l_3$ and $l_6$, which have the largest 
values in the second dimension and the first dimension,
respectively,
and construct a pattern $Q_s$ in 
Figure~\ref{fig:label-embedding}(3).
We can verify that 
(I) $Q_s$ is homomorphic
to none of the edges in Figure~\ref{fig:label-embedding}(1).
(II)  Since $l_3$ and $l_6$ have the largest 
values in the second dimension and the first dimension,
respectively,
the embedding of $Q_s$ is larger than embeddings
of all edges in Figure~\ref{fig:label-embedding}(1),
\ie the six edges in Figure~\ref{fig:label-embedding}(1)
can be homomorphic to $Q_s$,
which contradicts the fact that 
$l_1, \ldots, l_6$ are distinct.

\subsection*{A4. Proof of Theorem~\ref{thm-generalization-bound}}
We establish the Rademacher complexity  of \HFrame
by extending the proof of 
the Rademacher complexity for \GNNs~\cite{Gen-bound}.
Since \dual does not have any parameters, 
and works for any inputs, we only 
bound the Rademacher complexity of \HGIN.
We prove the bound in five steps:
(1) represent the computation of \HGIN
by a tree;
(2) quantify the impact of change 
of parameters on the embeddings;
(3) bound the changes in prediction probability;
(4) bound the covering numbers for \HGIN;
and (5) bound the Rademacher complexity
using the covering numbers.
We omit the details due to page limits.

\subsection*{A5. Complex queries}
\begin{figure}[th!]
	\vspace{-1.4ex}
	\centerline{\includegraphics[width=0.7\columnwidth]{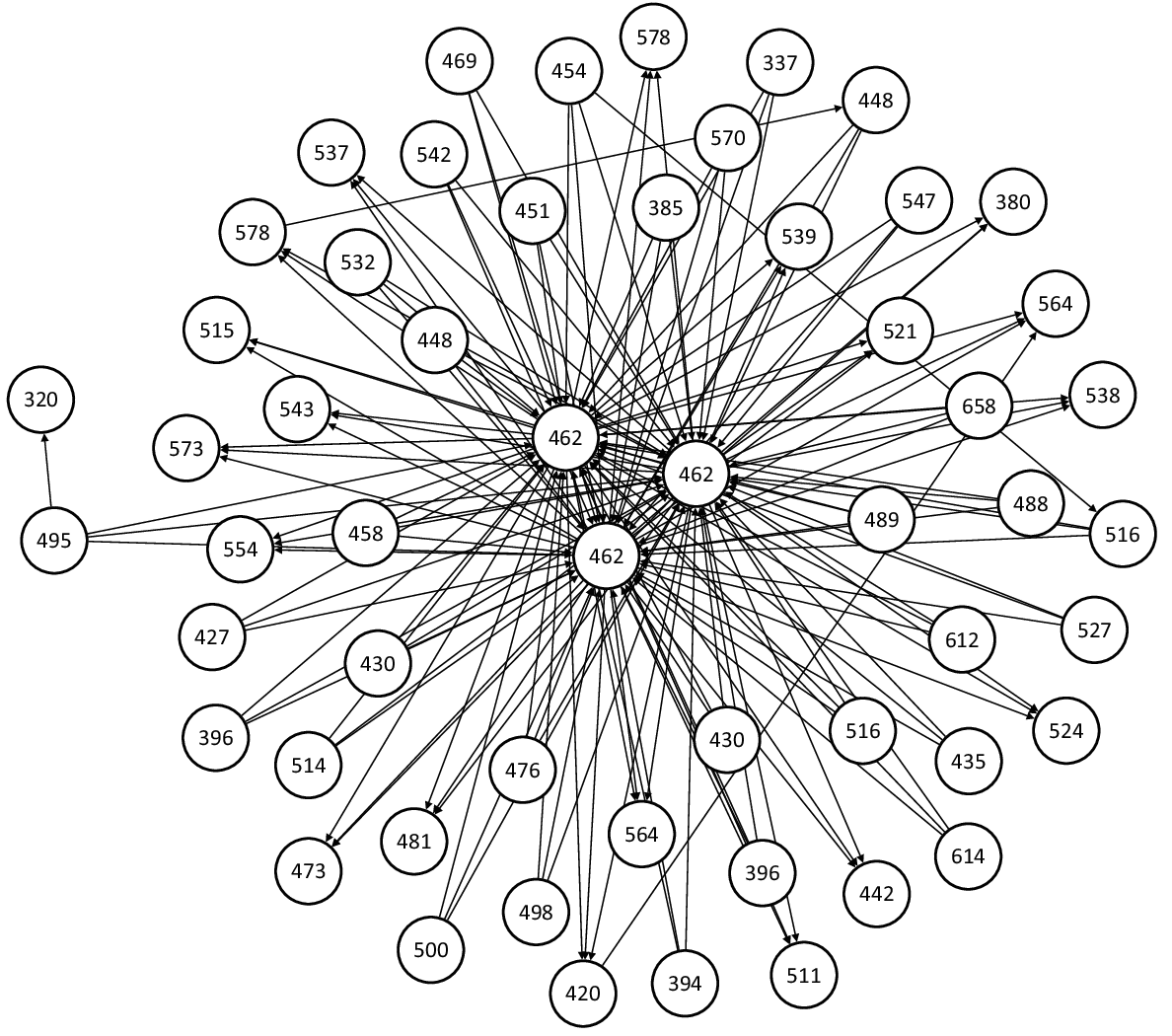}}
	\centering
	\vspace{-1.4ex}
	\caption{Example of complex query}
	\label{fig:query}
	\vspace{-1.4ex}
\end{figure}

For complex  queries  with multiple cycles
and a large number of vertices,
exact algorithms can be  slow 
and \HFrame is much faster than 
them. 
Given a  query with $55$ vertices (Figure~\ref{fig:query}) extracted from Citeseerx,
\kw{SIM}-\kw{TD}, \kw{PJ} and \kw{EJ} took
1.138s, 6.279s and over 30 seconds, respectively, 
to conduct the query over an extracted  subgraph with $1146$ vertices and $2176$ edges,
while \HFrame took only 0.074s.

\end{document}